\documentclass[10pt]{article} 
\usepackage[preprint]{tmlr}

\usepackage{amsmath,amsfonts,bm}

\def\eqref#1{equation~\ref{#1}}

\def\1{\bm{1}}

\DeclareMathAlphabet{\mathsfit}{\encodingdefault}{\sfdefault}{m}{sl}
\SetMathAlphabet{\mathsfit}{bold}{\encodingdefault}{\sfdefault}{bx}{n}

\usepackage[hyphens]{url}  
\usepackage{graphicx} 
\usepackage{algorithm}

\usepackage{newfloat}
\usepackage{listings}
\floatstyle{ruled}
\newfloat{listing}{tb}{lst}{}
\floatname{listing}{Listing}

\usepackage[utf8]{inputenc} 
\usepackage{booktabs}       
\usepackage{amsfonts}       
\usepackage{nicefrac}       
\usepackage{microtype}      
\usepackage{xcolor}         

\usepackage{adjustbox}

\usepackage{subcaption}
\usepackage{etoolbox}
\usepackage{verbatim}

\usepackage{mathrsfs}
\usepackage{bm}
\usepackage{algorithm}
\usepackage{algcompatible}
\usepackage[noend]{algpseudocode}
\usepackage{mathtools}
\usepackage{nccmath}
\usepackage{float}
\usepackage{multirow,color,graphicx}
\usepackage{subcaption}
\usepackage{caption}
\usepackage{xfrac,amsbsy}
\usepackage[titletoc]{appendix}
\usepackage{xspace}
\usepackage{savesym,verbatim}
\usepackage{paralist}

\usepackage{subscript}
\usepackage{tikz}
\usetikzlibrary{shapes.geometric}
\usetikzlibrary{shapes.misc}
\usetikzlibrary{decorations.pathreplacing}
\usetikzlibrary{patterns}
\usetikzlibrary{arrows,positioning,shapes}
\usetikzlibrary{arrows.meta}
\usepackage{picture}
\tikzset{cross/.style={cross out, draw=black, fill=none, minimum size=2*(#1-\pgflinewidth), inner sep=0pt, outer sep=0pt}, cross/.default={2pt}}
\usepackage{amsthm}
\usepackage{bbm}

\definecolor{ShadeColour}{RGB}{105, 105, 105}

\usepackage{environ}
\usepackage{esvect} 

\NewEnviron{boxcode2col}[4]{
    \begin{center}
   	 	\scalebox{#3}{
	   	 	\setlength\tabcolsep{4pt}
	   	 	\renewcommand{\arraystretch}{#4}
	   	 	\begin{tabular}{|p{#1}p{#2}|}   
    			\hline
				\BODY
   			 	[1ex]
    			\hline
    		\end{tabular}
    	}
    \end{center}
}

\NewEnviron{boxcode}[3]{
    \begin{center}
   	 	\scalebox{#2}{
	   	 	\setlength\tabcolsep{4pt}
	   	 	\renewcommand{\arraystretch}{#3}
    		\begin{tabular}{|p{#1}|}    
    			\hline
    			\vspace{0.1mm}    
				\BODY
    			\\\hline
    		\end{tabular}
    	}
    \end{center}
}

\newcommand{\textcode}[1]{{\fontfamily{cmtt}\selectfont #1}\xspace}

\newcommand{\taskspace}{\ensuremath{\mathbb{T}}}
\newcommand{\codespace}{\ensuremath{\mathbb{C}}}
\newcommand{\task}{\text{\textcode{T}}}
\newcommand{\code}{\text{\textcode{C}}}
\newcommand{\reducedcode}{\text{\textsc{RedCode}}}
\newcommand{\progression}{\ensuremath{\omega}}
\newcommand{\progressionset}{\ensuremath{\Omega}}

\newcommand{\FDissimilarity}{\ensuremath{\mathcal{F}_{\textnormal{diss}}^\mathbb{T}}}

\newcommand{\FTaskquality}{\ensuremath{\mathcal{F}_{\textnormal{qual}}^\mathbb{T}}}

\newcommand{\FComplexity}{\ensuremath{\mathcal{F}_{\textnormal{complex}}}}
\newcommand{\FCodecomplexity}{\ensuremath{\mathcal{F}_{\textnormal{complex}}^{\mathbb{C}}}}
\newcommand{\FTaskcomplexity}{\ensuremath{\mathcal{F}_{\textnormal{complex}}^{\mathbb{T}}}}
\newcommand{\FProgressioncomplexity}{\ensuremath{\mathcal{F}_{\textnormal{complex}}^{{\Omega}}}}

\newcommand{\DSLMove}{{\textcode{{move}}}}

\newcommand{\DSLTurnLeft}{\textcode{turnLeft}}

\newcommand{\DSLTurnRight}{\textcode{turnRight}}

\newcommand{\DSLPickMarker}{\textcode{pickMarker}}

\newcommand{\DSLPutMarker}{\textcode{putMarker}}
\newcommand{\DSLRepeat}{\textcode{\textsc{Repeat}}}
\newcommand{\DSLRepeatUntil}{\textcode{\textsc{RepeatUntil}}}
\newcommand{\DSLIf}{\textcode{\textsc{If}}}
\newcommand{\DSLIfElse}{\textcode{\textsc{IfElse}}}

\newcommand{\DSLWhile}{\textcode{\textsc{While}}}
\newcommand{\DSLRun}{\textcode{\textsc{Run}}}

\NewEnviron{boxcodecfg}[1]{%
\fbox{\begin{minipage}{#1}
  \begin{flalign*}
    \BODY
  \end{flalign*}
 \end{minipage}}
}

\newcommand{\defaultTraining}{\text{\textsc{Same-}\task\code}}

\newcommand{\default}{\text{\textsc{Default}}}
\newcommand{\nodecomp}{\text{\textsc{Same-}\task\code}}
\newcommand{\harddecomp}{\text{\textsc{Same-}\code}}
\newcommand{\expnew}{\text{\textsc{Crafted}-v1}}
\newcommand{\expopt}{\text{\textsc{Crafted}-v2}}
\newcommand{\human}{\text{\textsc{Crafted}}}
\newcommand{\noharddecomp}{\text{\textsc{Same}}}

\newcommand{\hocd}{\text{H}\textsubscript{08}}
\newcommand{\hocg}{\text{H}\textsubscript{16}}

\newcommand{\executiontrace}{\ensuremath{\Lambda^\textnormal{all}}}
\newcommand{\executiontracetask}{\ensuremath{\lambda_\tau}}
\newcommand{\executiontracecode}{\ensuremath{\lambda_\tau}}
\newcommand{\executiontracefiltered}{\ensuremath{\Lambda^\textnormal{filter}}}
\newcommand{\executiontracesymexec}{\ensuremath{\Lambda^\textnormal{SE}}}

\newcommand{\synsubtasks}{\text{\textsc{ProgresSyn}}}
\newcommand{\synsubtasksingle}{\text{\textsc{ProgresSyn}\textsuperscript{single}}}
\newcommand{\synsubtasksmulti}{\text{\textsc{ProgresSyn}\textsuperscript{grids}}}
\newcommand{\ours}{\text{\textsc{ProgresSyn}}}

\algdef{SE}[SUBALG]{Indent}{EndIndent}{}{\algorithmicend\ }%
\algtext*{Indent}
\algtext*{EndIndent}
\usepackage{footmisc}
\usepackage{hyperref}
\graphicspath {{figs/}}

\usepackage{url}

\newtoggle{MainSuppContent}
\settoggle{MainSuppContent}{true}

\newtoggle{MainContentOnly}
\settoggle{MainContentOnly}{false}

\newtoggle{SuppContentOnly}
\settoggle{SuppContentOnly}{false}

\title{Synthesizing a Progression of Subtasks \\for Block-Based Visual Programming Tasks}

\author{\name Alperen Tercan\thanks{Alperen Tercan did this work as part of an internship at the Max Planck Institute for Software Systems (MPI-SWS), Germany. Alperen Tercan led the implementation of the \ours{} algorithm and its evaluation with AI agents (neural program synthesizers); Ahana Ghosh led the evaluation with novice programmers via a user study.} \email atercan@mpi-sws.org \\
      \addr Max Planck Institute for Software Systems
      \AND
      \name Ahana Ghosh\footnotemark[1] \email gahana@mpi-sws.org \\
      \addr Max Planck Institute for Software Systems
      \AND
      \name Hasan Ferit Eniser \email hfeniser@mpi-sws.org\\
      \addr Max Planck Institute for Software Systems
      \AND
      \name Maria Christakis \email maria.christakis@tuwien.ac.at  \\
      \addr TU Wien
      \AND
      \name Adish Singla \email adishs@mpi-sws.org \\
      \addr Max Planck Institute for Software Systems
}

\def\month{12}  
\def\year{2022} 
\def\openreview{\url{https://openreview.net/forum?id=XXXX}} 

\begin{document}

\maketitle

\begin{abstract}
\looseness-1 Block-based visual programming environments play an increasingly important role in introducing computing concepts to K-$12$ students. In recent years, they have also gained popularity in neuro-symbolic AI, serving as a benchmark to evaluate general problem-solving and logical reasoning skills. The open-ended and conceptual nature of these visual programming tasks make them challenging, both for state-of-the-art AI agents as well as for novice programmers. A natural approach to providing assistance for problem-solving is breaking down a complex task into a progression of simpler subtasks; however, this is not trivial given that the solution codes are typically nested and have non-linear execution behavior. In this paper, we formalize the problem of synthesizing such a progression for a given reference block-based visual programming task. We propose a novel synthesis algorithm that generates a progression of subtasks that are high-quality, well-spaced in terms of their complexity, and solving this progression leads to solving the reference task. We show the utility of our synthesis algorithm in improving the efficacy of AI agents (in this case, neural program synthesizers) for solving tasks in the \emph{Karel programming environment}~\citep{pattis1995karel}. Then, we conduct a user study to demonstrate that our synthesized progression of subtasks can assist a novice programmer in solving tasks in the \emph{Hour of Code: Maze Challenge}~\citep{hourofcode_maze} by \emph{Code.org}~\citep{codeorg}. 
\end{abstract}


\section{Introduction}\label{sec.intro}
The emergence of block-based visual programming platforms has made coding more accessible and appealing to novice programmers, including K-$12$ students. Led by the success of programming environments like \emph{Scratch}~\citep{DBLP:journals/cacm/ResnickMMREBMRSSK09} and \emph{Karel}~\citep{pattis1995karel}, initiatives like \emph{Hour of Code}~\citep{hourofcode} by \emph{Code.org}~\citep{codeorg} and online platforms like \emph{CodeHS.com}~\citep{codehscom}, block-based programming has become an integral part of introductory computer science education. Importantly, in contrast to typical text-based programming, block-based visual programming reduces the burden of learning syntax and puts direct emphasis on fostering computational thinking and general problem-solving~\citep{DBLP:conf/acmidc/WeintropW15,Price2017PositionPB, Price2015}. This unique aspect, in turn, also makes block-based visual programming environments an interesting benchmark for neuro-symbolic AI, in particular, to evaluate agents' problem-solving and logical reasoning skills~\citep{DBLP:conf/nips/SchusterKPK21,DBLP:conf/nips/Puri0JZDZD0CDTB21,DBLP:journals/corr/abs-2203-07814}. 
%


\begin{figure}[!t]
\centering
%
    \begin{subfigure}[b]{0.3\textwidth}
    \centering{
        \includegraphics[height=3.2cm]{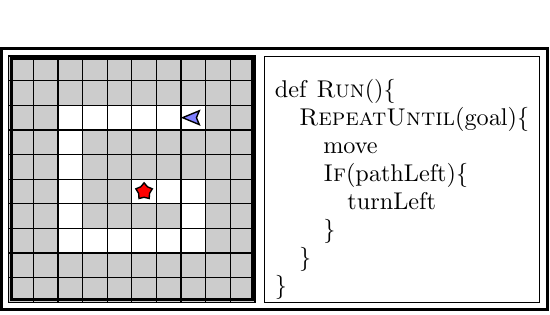}
    }
    \vspace{-2mm}    
    \caption{Reference task: $(\task^\textnormal{ref}, \code^{\textnormal{ref},\ast})$}
    \label{fig:intro.hoc.og}
    \end{subfigure}
    \\
    \begin{subfigure}[b]{0.35\textwidth}
    \centering{
        \includegraphics[height=3.2cm]{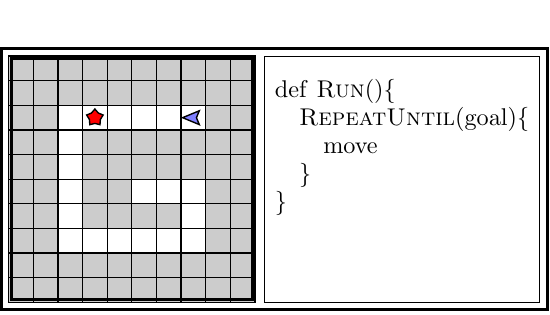}
    }
    \caption{Subtask 1: $(\task^{1}, \code^{1,\ast})$}
    \label{fig:intro.hoc.subtask1}
    \end{subfigure}
    \begin{subfigure}[b]{0.35\textwidth}
    \centering{
        \includegraphics[height=3.2cm]{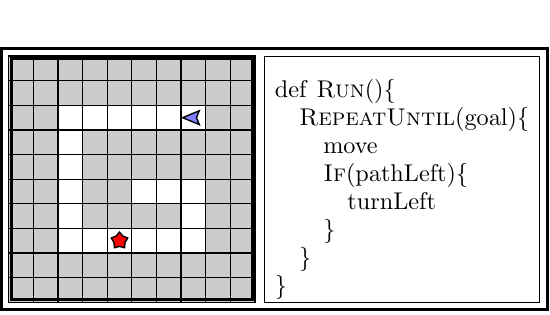}
    }
    \caption{Subtask 2: $(\task^{2}, \code^{2,\ast})$}
    \label{fig:intro.hoc.subtask2}
    \end{subfigure}
    \hspace{-5mm}
    \begin{subfigure}[b]{0.275\textwidth}
    \centering{
    \includegraphics[height=2.6cm]{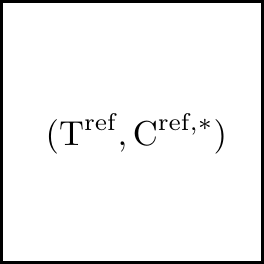}
    }
    \vspace{1.mm}     
    \caption{Subtask 3: $(\task^{3}, \code^{3,\ast})$}
    \label{fig:intro.hoc.subtask3}
    \end{subfigure}
    \vspace{3mm}
    \caption{Illustration of our synthesis algorithm on Maze$16$ task from the \emph{Hour of Code: Maze Challenge}~\citep{hourofcode_maze} by \emph{Code.org}~\citep{codeorg}. \textbf{(a)} shows visual grid $\task^\textnormal{ref}_\textnormal{vis}$ of reference task $\task^\textnormal{ref}$ and its solution code $\code^{\textnormal{ref},\ast}$ which are provided as input to our synthesis algorithm. When solving this task, one needs to write a code that upon execution would navigate the ``avatar'' (purple dart) to the ``goal'' (red star) in the visual grid. Additionally, the maximum number of allowed code blocks in the solution code is $\task^\textnormal{ref}_\textnormal{size} = 4$ and allowed types of blocks are $\task^\textnormal{ref}_\textnormal{store}$ = \{\DSLRepeatUntil{},\DSLIf{}, \DSLMove{},\DSLTurnLeft{},\DSLTurnRight{}\}. \textbf{(b), (c)}, and \textbf{(d)} show the progression of $K=3$ subtasks for the reference task synthesized by our algorithm~\synsubtasks. As can be seen, the subtasks are well-spaced w.r.t. their complexity and the visual grids are minimal modifications of the visual grid $\task^\textnormal{ref}_\textnormal{vis}$. 
    }
    \vspace{-1.5mm}
    \label{fig:intro.hoc}

\end{figure}

%


\begin{figure*}[!t]
    \centering
    %
    \begin{subfigure}[b]{1.\textwidth}
    \centering{
    \includegraphics[height=4.55cm]{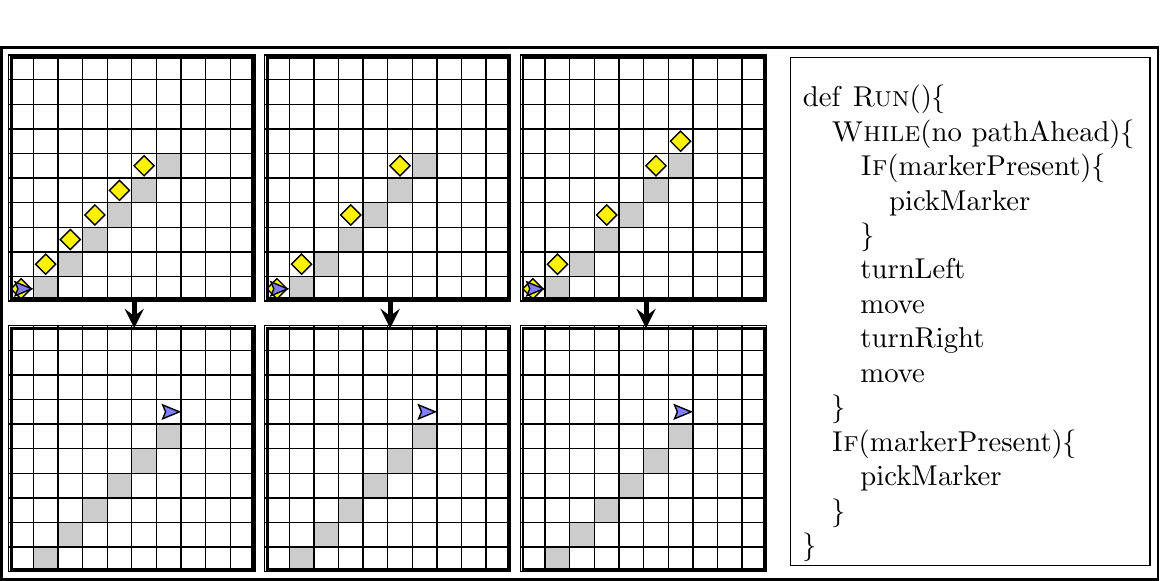}
    }
    \vspace{1.5mm}  
    \caption{Reference task: $(\task^\textnormal{ref}, \code^{\textnormal{ref},\ast})$}
    \label{fig:intro.multigrid.og}
    \end{subfigure}
    \\
    \begin{subfigure}[b]{0.3\textwidth}
    \centering{
    \includegraphics[height=4.55cm]{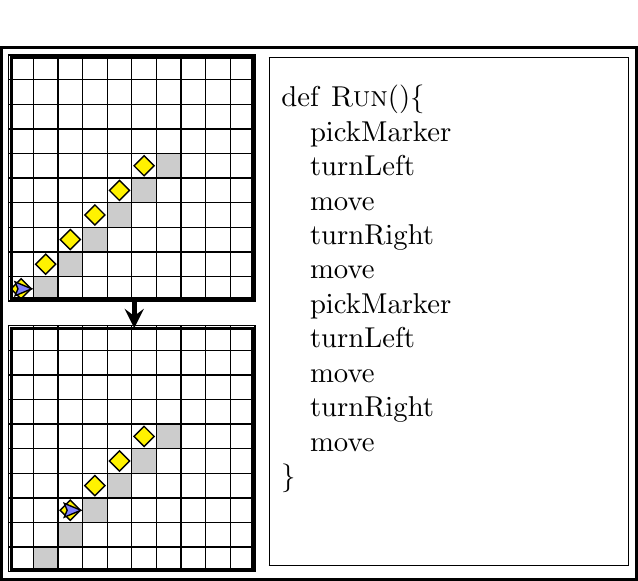}
    }
    \caption{Subtask 1: $(\task^{1}, \code^{1,\ast})$}
    \end{subfigure}
    \begin{subfigure}[b]{0.3\textwidth}
    \centering{
        \includegraphics[height=4.55cm]{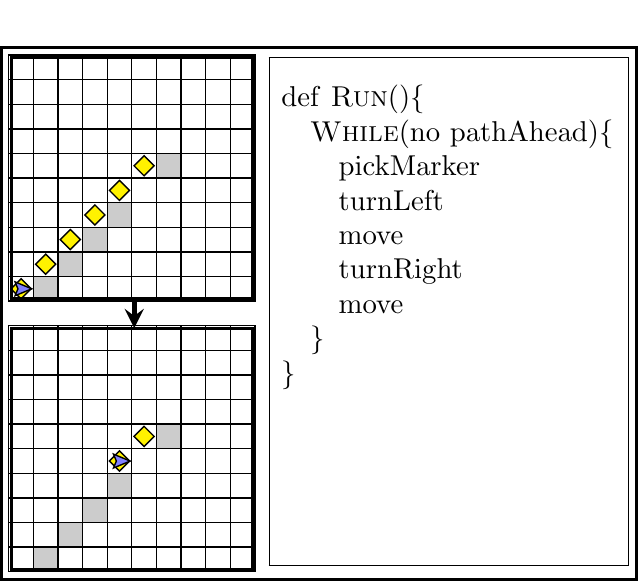}
    }
    \caption{Subtask 2: $(\task^{2}, \code^{2,\ast})$}
    \end{subfigure}
    \begin{subfigure}[b]{0.3\textwidth}
        \centering{
            \includegraphics[height=4.55cm]{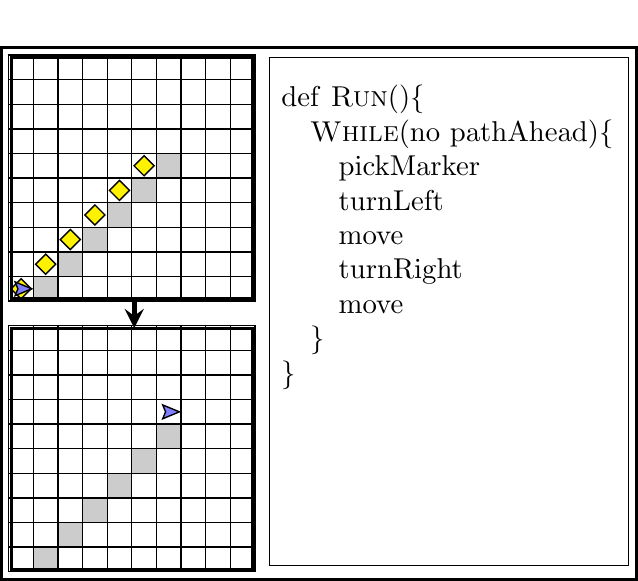}
    }
    \caption{Subtask 3: $(\task^{3}, \code^{3,\ast})$}
    \end{subfigure}
    \\
    \begin{subfigure}[b]{0.55\textwidth}
        \centering{
        \includegraphics[height=4.55cm]{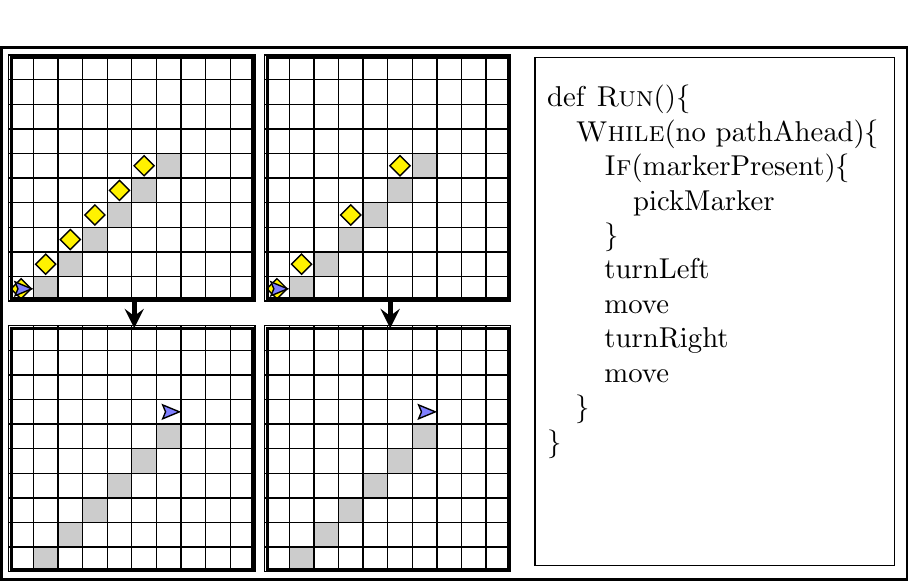}
    }
    \vspace{1.5mm}
    \caption{Subtask 4: $(\task^{4}, \code^{4,\ast})$}
    \end{subfigure}
    \begin{subfigure}[b]{0.35\textwidth}
        \centering{
        \includegraphics[height=4.15cm]{figs/appendix-userstudy/empty_taskcode.pdf}
    }
    \vspace{1.5mm}   
    \caption{Subtask 5: $(\task^{5}, \code^{5,\ast})$}
    \end{subfigure}
    \caption{\looseness-1Illustration of our synthesis algorithm on an adaptation of \textsc{Stairway} task from \emph{CodeHS.com} \citep{codehscom} based on \emph{Karel programming environment} \citep{pattis1995karel}. Compared to the task shown in Figure~\ref{fig:intro.hoc}, Karel tasks are more complex and include multiple I/O pairs. \textbf{(a)} shows the three I/O pairs of reference task $\task^\textnormal{ref}$ and its solution code $\code^{\textnormal{ref},\ast}$, provided as input to our synthesis algorithm. When solving the task, the goal is to write code that upon execution, converts pregrid (upper grid of a visual I/O pair) to its corresponding postgrid (lower grid of a visual I/O pair, and connected to the pregrid by an arrow) for all I/O pairs. The visual grids additionally have yellow diamond ``markers'' which can be put or picked from a grid cell using the \DSLPutMarker{} and \DSLPickMarker{} code blocks respectively.
    \textbf{(b--f)} show the progression of $K=5$ subtasks for the reference task synthesized by our algorithm~\synsubtasks. In particular, the subtasks in (b-d) are synthesized from the first I/O pair in (a) and the subtasks in (e-f) gradually bring in the other two I/O pairs of (a). 
    }
    %
    \label{fig:intro.multigrid}
    \vspace{-3.5mm}
\end{figure*}

%

Programming tasks on these platforms are open-ended and require multi-step deductive reasoning to solve. As a result, novices may struggle when solving these tasks, leading to low success rate in finding a correct solution~\citep{DBLP:conf/lats/PiechSHG15,DBLP:conf/aied/PriceZB17,DBLP:conf/aaai/WuMGP19,edm20-zero-shot,aied21_popquiz}. Similarly, solving these tasks can also be challenging for state-of-the-art AI agents (neural program synthesizers) trained using reinforcement learning methods~\citep{DBLP:journals/corr/abs-2107-08828}; as observed in~\citet{DBLP:conf/iclr/BunelHDSK18}, the agents' performance decreased drastically with increased nesting in the solution codes. A natural way to deal with task complexity is to ``break down a task'' into subtasks---this framework of subtasking has proven to be effective in several domains~\citep{DBLP:conf/icer/DeckerMM19,DBLP:conf/icer/MorrisonMG15}, including geometry proof problems~\citep{DBLP:journals/hhci/McKendree90}, Parson's coding problems~\citep{DBLP:conf/sigcse/MorrisonMEG16}, and robotics~\citep{bakker2004hierarchical, ding2014autonomic}. Inspired by this, we seek to develop algorithms that can synthesize a progression of subtasks for block-based visual programming tasks.

However, automatically synthesizing such a progression for block-based visual programming domains is non-trivial. Codes have nested structures, their execution behaviour on task's visual grid is ``non-linear''; hence, existing subtasking techniques (in domains such as path-navigation or robotics) relying on well-defined ``linear'' behaviors do not apply to our setting~\citep{bakker2004hierarchical, ding2014autonomic}.
Also, the space of visual grids and codes is potentially unbounded, and the mapping between these spaces is highly discontinuous (i.e., a small modification can make the task invalid); hence, techniques relying on exhaustive enumeration to find valid/high-quality subtasks are intractable~\citep{DBLP:conf/nips/AhmedCEFGRS20, DBLP:conf/ijcai/PolozovOSZGP15}.

\subsection{Our Approach and Main Contributions}
We begin by formalizing our objective of synthesizing a progression of programming subtasks. Our proposed synthesis methodology overcomes the key challenges discussed above by reasoning about the execution behavior of the provided solution code on the reference task's visual grids. 
As concrete examples, consider reference tasks in Figures~\ref{fig:intro.hoc.og}~and~\ref{fig:intro.multigrid.og} from two popular platforms. Given such a reference task along with a solution code as input, we seek to synthesize a progression of subtasks with the following properties: (a) each subtask is a high-quality programming task by itself; (b) the complexity of solving a subtask in the progression increases gradually, i.e., the subtasks are well-spaced w.r.t. their complexity; (c) solving the progression would help in increasing success rate of solving the reference task.

Our main contributions and results are as follows: (I) We formalize the objective of synthesizing a progression of subtasks for block-based visual programming tasks and present a novel synthesis algorithm, \synsubtasks{} (Sections~\ref{sec.problemsetup}~and~\ref{sec.algorithm}); (II) We showcase the utility of \synsubtasks{} towards assisting the problem-solving process of AI agents (neural program synthesizers) and of novice programmers via an online study (Sections~\ref{sec.simulations}~and~\ref{sec.userstudy});  (III) We publicly share the web app used in the study\footnote{Link:\url{https://www.teaching-blocks-subtasks.cc}} and the implementation of \synsubtasks{}\footnote{Link:\url{https://github.com/machine-teaching-group/ProgresSyn}\label{fn:github-repo}} to facilitate future research (Sections~\ref{sec.algorithm} and \ref{sec.userstudy.setup}).


\subsection{Related Work}

\paragraph{Subtasks in programming education.} Prior work in developing subtasks for block-based programming tasks requires access to resources such as expert labels or students’ historical data on these platforms \citep{DBLP:conf/iticse/MargulieuxMD19,DBLP:journals/jeric/MargulieuxMFR20,DBLP:conf/edm/Marwan0MCBP21}. For instance, \citet{DBLP:conf/edm/Marwan0MCBP21} uses unsupervised clustering techniques to automatically detect common code patterns in student data, followed by hierarchical clustering methods to combine frequently co-occurring code patterns into subtasks. Our approach is different in that we seek to automate the generation of subtasks without access to any prior data. 

\looseness-1\paragraph{Neural program synthesis (NPS).} 
In recent years, a number of neural models have been proposed which learn to synthesize solution codes for text-based and visual block-based programming tasks~\citep{DBLP:conf/iclr/BalogGBNT17,DBLP:journals/corr/abs-2108-07732,DBLP:conf/icml/DevlinUBSMK17,DBLP:conf/iclr/ChenLS19}. However, these models are data hungry, which has led to a significant amount of work in creating synthetic datasets for training these models~\citep{DBLP:conf/nips/AhmedCEFGRS20,DBLP:conf/iclr/LaichBV20,DBLP:conf/iclr/BunelHDSK18,DBLP:journals/corr/abs-2003-10485,edm22-student-synthesis}. \ours{} complements this line of work by decomposing existing tasks into subtasks -- these subtasks when augmented with the original dataset would further increase the diversity of the dataset. We validate this property of our synthesized subtasks in Section~\ref{sec.simulations}. An alternate approach to training neural program synthesizers (as in the work of \cite{DBLP:conf/iclr/ChenLS19}), would be to dynamically break down the task into intermediate steps during a training episode and reason about partial codes. However, this approach does not synthesize standalone subtasks, which is the focus of our study.

%


\section{Problem Setup}\label{sec.problemsetup}
In this section, we introduce definitions and formalize our objective.
We have provided a detailed table of notations in Appendix~\ref{appendix.sec.problemsetup}.

\subsection{Preliminaries}\label{sec:problemstatement.prelims}

\looseness-1\paragraph{The space of tasks and codes.} We define a task \task~as a tuple $(\task_\textnormal{n},\task_\textnormal{vis}, \task_\textnormal{store}, \task_\textnormal{size})$ where $\task_\textnormal{n}$ denotes the number of visual grids, $\task_\textnormal{vis} := \{\task_{\textnormal{vis,}i}\}_{i=1,\ldots,\task_\textnormal{n}}$ denotes the set of $\task_\textnormal{n}$ visual grids, $\task_\textnormal{store}$ denotes the types of code blocks available, and $\task_\textnormal{size}$ denotes the maximum number of code blocks allowed in the solution code. 
We denote the space of tasks as \taskspace. We also define a few important properties of tasks in this space through functions that can be instantiated as desired: (i) $\FTaskcomplexity: \taskspace \rightarrow \mathbb{R}$ measures the complexity of solving a task; (ii)$\FTaskquality: \taskspace \rightarrow \mathbb{R}$ measures the general quality of a task~\citep{DBLP:conf/nips/AhmedCEFGRS20}; (iii) $\FDissimilarity: \taskspace\times\taskspace \rightarrow \mathbb{R}$ captures the dissimilarity between any two tasks in \taskspace. For instance, given the visual nature of the tasks, a possible dissimilarity metric can be the hamming-distance between their visual grids~\citep{DBLP:conf/nips/0002FS12, DBLP:conf/nips/AhmedCEFGRS20}. The domain specific language (DSL) of the programming environment defines the space of codes \codespace~\citep{DBLP:conf/nips/AhmedCEFGRS20, DBLP:conf/iclr/BunelHDSK18}. A code $\code \in \codespace$ is characterized by the following attributes: $\code_\textnormal{depth}$ measures the depth of the abstract syntax tree (AST) of \code, $\code_\textnormal{size}$ is the number of code blocks in \code, and $\code_\textnormal{blocks}$ is the types of code blocks in \code. We also define a code complexity measure using function $\FCodecomplexity$. Typically, in block-based programming environments, complexity of a code depends on the depth and size of its AST~\citep{DBLP:conf/lats/PiechSHG15,aied21_popquiz}. Motivated by this, we define $\FCodecomplexity = \kappa * \code_\textnormal{depth} + \code_\textnormal{size}$, where $\kappa \in \mathbb{N}$. For instance, empty code \{\DSLRun{}\} has complexity $\kappa*1 + 0$, and code \{\DSLRun{}\{\DSLRepeat{}(\textcode{4})\{\text{\DSLMove{}\}}\}\} has complexity $\kappa*2 + 2$.

\paragraph{Solution code of a task.} We define $\code \in \codespace$ as a solution code for a task $\task \in \taskspace$ if all of the following conditions hold: execution of \code~solves all the $\task_\textnormal{n}$ visual grids of \task, $\code_\textnormal{size} \leq \task_\textnormal{size}$, and $\code_\textnormal{blocks} \subseteq \task_\textnormal{store}$. We denote a specific solution code of a task \task~as $\code^{\task,\ast}$; for a task $\task^{\textnormal{id}}$, we denote its solution code as $\code^{\textnormal{id},\ast}$ for brevity. We denote $\codespace_{\task}$ as the set of all solution codes of \task. Using the notion of solution code of a task, we can specify our complexity measure of a task in this domain. In particular, we define $\FTaskcomplexity(\task) = \min_{\code \in \codespace_{\task}} \FCodecomplexity(\code)$.\footnote{In  block-based visual programming domains considered here, the set of solution codes for a given reference task $\codespace_\task$ is typically small in size. However, if this is not the case, we can define $\codespace_\task$ as a small set of representative solution codes for the reference task.}

\subsection{Objective}\label{sec:problemstatement.obj}
Our goal is to synthesize a progression of subtasks for a given reference task, such that solving this progression increases the success rate of solving the reference task. We use the increased success rate as a proxy for measuring the helpfulness of our synthesized progression.
Next, we introduce the notion of progression of subtasks and its complexity, and then formalize our synthesis objective.

\paragraph{Progression of subtasks.} For a reference task $\task^\textnormal{ref}$, its solution code $\code^{\textnormal{ref},\ast}$, and a fixed budget $K$, we denote a progression of subtasks for $\task^\textnormal{ref}$ as a sequence $\progression(\task^\textnormal{ref}, \code^{\textnormal{ref},\ast}, K) := ((\task^k, \code^{k,\ast}))_{k=1,2,\ldots,K}$ where the following holds $\forall k$: (a) $\code^{k,\ast}$ is the solution code of $\task^{k}$; (b) $\task_\textnormal{n}^{k} \leq \task_\textnormal{n}^\textnormal{ref}$; (c) $\task_\textnormal{store}^{k} \subseteq \task^\textnormal{ref}_\textnormal{store}$. 
We also have $\task^K = \task^\textnormal{ref}$ and $\code^{K,\ast} = \code^{\textnormal{ref},\ast}$. We denote the set of all such progressions of $K$-subtasks as $\progressionset({\task^\textnormal{ref}}, \code^{\textnormal{ref},\ast}, K)$. 

\paragraph{Complexity of a progression of subtasks.} 
We capture the complexity of a progression of subtasks using the function \FProgressioncomplexity.  Specifically, $\FProgressioncomplexity(\progression; \task^\textnormal{ref}, \code^{\textnormal{ref},\ast}, K)$ for a given reference task $\task^\textnormal{ref}$ captures the worst case complexity jump in the solution codes of subtasks of \progression. More formally:
\begin{align*}
\small\FProgressioncomplexity(\progression; \task^\textnormal{ref}, \code^{\textnormal{ref},\ast},K) = 
\max_{k \in \{1,\ldots,K\}} \small\Big\{\min_{k'\in \{0,\ldots,k-1\}} \big\{\FCodecomplexity(\code^{k,\ast}) - \FCodecomplexity(\code^{k',\ast})\big\}\Big\}
\tag{1}
\label{eq:problemstatement.0}
\end{align*}
where $\code^{k,\ast}$/$\code^{k',\ast}$ denote solution codes of subtasks $k$/$k'$ in \progression, and $\code^{0,\ast}$ denotes empty code \{\DSLRun{}\}.

\paragraph{Our synthesis objective.} Our objective is to synthesize a progression of $K$ subtasks for a given reference task $\task^\textnormal{ref}$ with minimal complexity w.r.t. Equation~\ref{eq:problemstatement.0}. Our formalism is based on the intuition that lowering complexity reduces the cognitive load of solving the progression, while still being helpful in assisting problem-solving of the reference task~\citep{DBLP:journals/hhci/McKendree90}.
More formally, we seek to generate a progression of subtasks based on the following:
\begin{align*}
\text{Minimize}_{\progression \in \progressionset(\task^\textnormal{ref},\code^{\textnormal{ref},\ast}, K)} \FProgressioncomplexity(\progression; \task^\textnormal{ref}, \code^{\textnormal{ref},\ast}, K).
\tag{2}
\label{eq:problemstatement.1}
\end{align*}

In addition, we seek to have the following three desirable properties in our synthesized progression of subtasks. First, subtasks should be of high quality, i.e., $\text{maximize}~\Sigma_{k \in \{1,\ldots,K\}} \FTaskquality(\task^k)$. Second, visual grids corresponding to each of the subtasks should be minimal modifications of the visual grids of the reference task, i.e., $\text{minimize}~\Sigma_{k\in\{1,\ldots,K\}} \FDissimilarity(\task^k, \task^\textnormal{ref})$. This is based on the intuition that minimal switch in visual context is better for problem-solving~\citep{burnett1995visual, kiel2009reducing}; see Section~\ref{sec.userstudy}. Third, subtasks in the progression should generally be diverse. In our implementation, we use these three properties for tie-breaking when solving Equation~\ref{eq:problemstatement.1}.\footnote{In Equation~\ref{eq:problemstatement.1}, we have defined the set of progressions \progressionset{} w.r.t a single solution code of the reference task; however, \progressionset{} could be extended to account for multiple solution codes.}
%



\begin{figure*}[t!]
\centering
    \begin{subfigure}[b]{1\textwidth}
        \centering{
        \includegraphics[height=3.24cm]{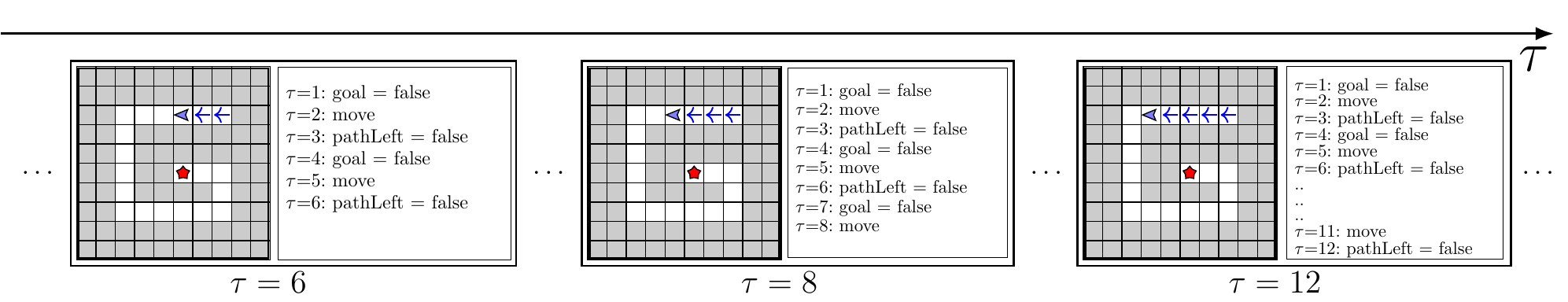}
        }
    \vspace{-2mm}
    \caption{Stage 1: Execution trace of $\code^{\textnormal{ref},\ast}$ on $\task^\textnormal{ref}_{\textnormal{vis,}1}$ denoted as \executiontrace}
    \label{fig:model2.hoc.stage1}
    \end{subfigure}
\\
\vspace{-2.4mm}
    \begin{subfigure}[b]{1\textwidth}
        \centering{
        \includegraphics[height=3.24cm]{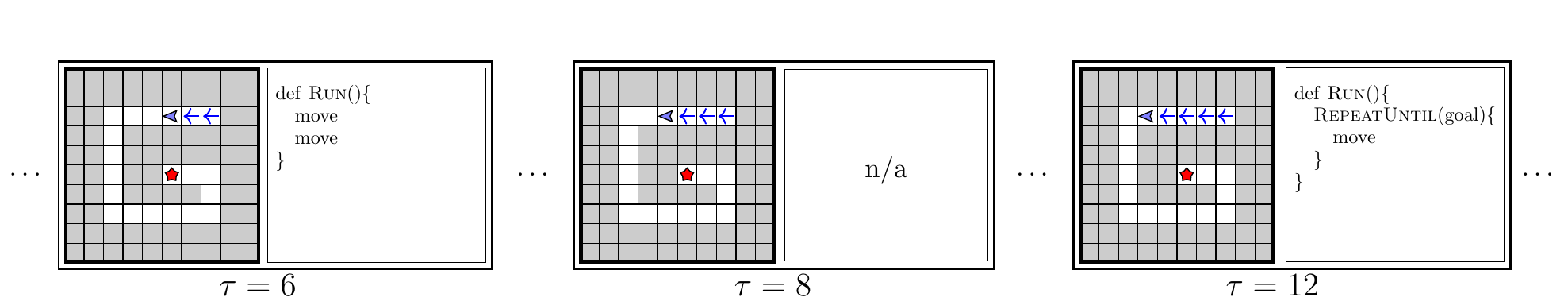}
        }
    \vspace{-2mm}
    \caption{Stage 2: Post-processing \executiontrace~based on code validity and code quality to obtain \executiontracefiltered}
    \label{fig:model2.hoc.stage2}
    \end{subfigure}
\\
\vspace{-2.4mm}
    \begin{subfigure}[b]{1\textwidth}
        \centering{
        \includegraphics[height=3.24cm]{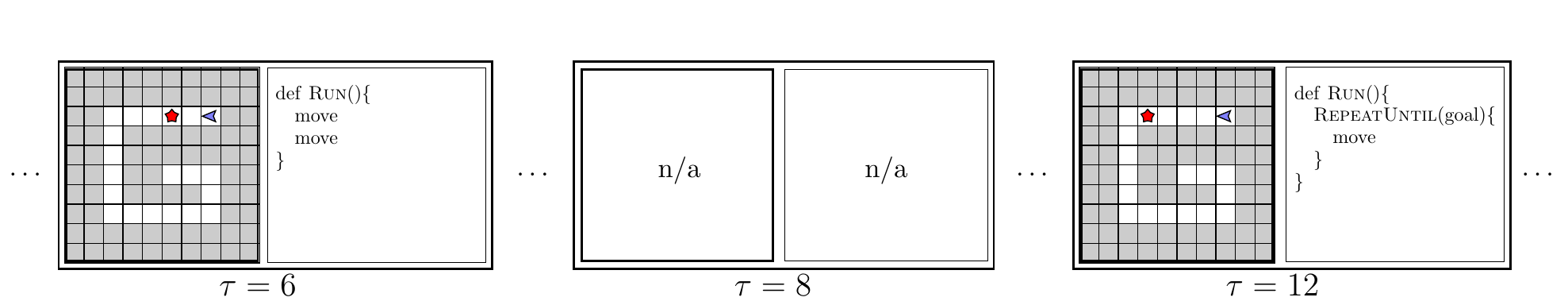}
        }
    \vspace{-2mm}
    \caption{Stage 3: Modifying grids in \executiontracefiltered~via symbolic execution to obtain \executiontracesymexec}
    \label{fig:model2.hoc.stage3}
    \end{subfigure}
\caption{Illustration of three stages of \synsubtasksingle~on reference task shown in Figure~\ref{fig:intro.hoc.og}. ``n/a'' denotes invalid codes and tasks. In Stage 4, we select a progression of $K'=3$ subtasks shown in Figures~\ref{fig:intro.hoc.subtask1}, \ref{fig:intro.hoc.subtask2}, and \ref{fig:intro.hoc.subtask3}. See the details in Section~\ref{sec.algorithm.singlegrid}.
}
\label{fig:model2.hoc}
\end{figure*}
%

%
\section{Our Synthesis Algorithm}\label{sec.algorithm}
\looseness-1In this section, we present our algorithm, \synsubtasks, for synthesizing a progression of subtasks for a reference task and its solution code $(\task^\textnormal{ref}, \code^{\textnormal{ref},\ast})$. \synsubtasks{} builds up on two procedures: \synsubtasksmulti{} for gradually introducing visual grids as subtasks and \synsubtasksingle{} for synthesizing subtasks for a task with a single visual grid.\footnote{While our proposed algorithm is designed for block-based visual programming environments, it could potentially be extended to other domains (such as text-based programming environments) if the following elements are redesigned: (a) task synthesis for programs in the domain using symbolic execution; (b) task complexity; (c) task dissimilarity.}

\subsection{\synsubtasksmulti}\label{sec.algorithm.multigrid}

\looseness-1Existing programming platforms such as \emph{HackerRank}~\citep{hackerrank}, \emph{Codeforces}~\citep{codeforces} and initiatives such as Code Hunt~\citep{bishop2015code} have test cases to guide the process of programming. 
In block-based visual programming, visual grids of a task are equivalent to test cases that a solution code must satisfy. We begin by describing how to construct subtasks from subsets of visual grids of a reference task and then present a procedure that generates a progression by gradually introducing the visual grids.

\looseness-1\paragraph{Generating a progression of subtasks.}
For a task \task~ and a subset of visual grids $G \subseteq \task_\textnormal{vis}$, we define a \emph{code reduction} $\hat{\code}$~as the code that solves $G$ and is obtained by removing inactive branches of $\code^{\task,\ast}$ during program execution~\citep{DBLP:conf/sigsoft/FerlesWCD17, DBLP:journals/ipl/KorelL88}. We denote a code reduction as $\hat{\code} := \reducedcode(G; \task,\code^{\task,\ast})$. Using these elements, we define a subtask of \task~as $\hat{\task} := (\hat{\task}_\textnormal{n}=|G|, \hat{\task}_{\textnormal{vis}}=G, \hat{\task}_\textnormal{store}=\hat{\code}_\textnormal{blocks}, \hat{\task}_\textnormal{size}=\hat{\code}_\textnormal{size})$. Using these subtasks, we develop a procedure \synsubtasksmulti~to generate our progression of subtasks. Before describing our procedure, we introduce some notation: $\Sigma^{n}$ denotes the collection of all permutations of the set \{1,2,\ldots,n\}; $\sigma \in \Sigma^n$ refers to one such permutation in the collection; $\sigma_i$ refers to the $i$-th element of $\sigma$. Our procedure begins by generating a progression of subtasks for a given $\sigma \in \Sigma^{\task^\textnormal{ref}_\textnormal{n}}$, denoted as $\progression^\sigma$. In $\progression^\sigma$, we define the $k$-th subtask with visual grids $G = \{\task^\textnormal{ref}_{\textnormal{vis},\sigma_i}\}_{i=1,\ldots,k}$ and solution code $\code = \reducedcode(G; \task^\textnormal{ref}, \code^\textnormal{ref})$ as $\task := (|G|, G, \code_\textnormal{blocks}, \code_\textnormal{size})$. Our procedure then generates a set of progressions of subtasks as $\progressionset(\task^\textnormal{ref}, \code^{\textnormal{ref},\ast}, \task^\textnormal{ref}_\textnormal{n}) := \{\progression^\sigma\}_{\sigma \in \Sigma^{\task^\textnormal{ref}_\textnormal{n}}}$. Using $\progressionset(\task^\textnormal{ref}, \code^{\textnormal{ref},\ast}, \task^\textnormal{ref}_\textnormal{n})$, we obtain our optimized progression of subtasks for $\task^\textnormal{ref}$ based on Equation~\ref{eq:problemstatement.1}. When $\task^\textnormal{ref}_\textnormal{n}$ is small, we can optimize for Equation~\ref{eq:problemstatement.1} by enumerating all possible elements from \progressionset. When $\task^\textnormal{ref}_\textnormal{n}$ is large, we can use greedy optimization strategies; we present further details in Appendix \ref{appendix.model.grids}.

\looseness-1\paragraph{Need for more fine-grained subtasks.} However, generating a progression of subtasks for reference task $\task^\textnormal{ref}$ using only subsets of its visual grids is limiting because of the following reasons: (a) When $\task^\textnormal{ref}_\textnormal{n} =1$, our procedure would return $\task^\textnormal{ref}$ as the only subtask which would not serve our intended purpose of reducing the complexity of the reference task; (b) Reference tasks with $\task^\textnormal{ref}_\textnormal{n} > 1$ could have all its visual grids requiring high code coverage of $\code^{\textnormal{ref},\ast}$, resulting in high complexity of all reduced codes and hence high complexity of all subtasks. To overcome these limitations, there is a need for a more fine-grained procedure to synthesize subtasks as discussed next.
%


%
\subsection{\synsubtasksingle}
\label{sec.algorithm.singlegrid}

\looseness-1Next, we describe a procedure \synsubtasksingle~for obtaining subtasks for a single-grid task $\task^\textnormal{ref}$ (i.e., $\task^\textnormal{ref}_\textnormal{n}=1$) and its solution code $\code^{\textnormal{ref},\ast}$. Our procedure first ``linearizes'' $\code^{\textnormal{ref},\ast}$ by obtaining its execution trace on $\task^\textnormal{ref}_\textnormal{vis,1}$. This makes it easier to segment and generate well-spaced codes for the progression of subtasks. Then, we process elements of the trace to generate codes which serve as solution codes of the subtasks. Next, we synthesize visual grids from the solution codes via minimal modifications w.r.t. to $\task^\textnormal{ref}_\textnormal{vis,1}$ using symbolic execution. Finally, we select a sequence of $K'$ subtasks to obtain our progression. These four stages are described below.
\looseness-1\paragraph{Stage 1: Execution trace on the single grid (Figure~\ref{fig:model2.hoc.stage1}).} 
This stage ``linearizes'' the solution code $\code^{\textnormal{ref},\ast}$ by obtaining the full execution trace of this code on single visual grid $\task^\textnormal{ref}_{\textnormal{vis},1}$. We denote the execution trace as the sequence $\executiontrace(\task^\textnormal{ref}, \code^{\textnormal{ref},\ast}) := ((\executiontracetask^{\task^\textnormal{ref}}, \executiontracecode^{\code^{\textnormal{ref},\ast}}))_{\tau=1,\ldots,M^\textnormal{all}}$ where $M^\textnormal{all}$ is the total steps in the execution, $\executiontracecode^{\code^{\textnormal{ref},\ast}}$ is the sequence of code commands executed till step $\tau$, and $\executiontracetask^{\task^\textnormal{ref}}$ is the state of $\task^\textnormal{ref}_{\textnormal{vis},1}$ at time-step $\tau$ after $\executiontracecode^{\code^{\textnormal{ref},\ast}}$ is executed. %

\looseness-1\paragraph{Stage 2: Post-processing the trace based on code validity/quality (Figure~\ref{fig:model2.hoc.stage2}).}
This stage filters the execution trace and generates potential solution codes of the subtasks. We begin by filtering those elements of the trace whose code commands lead to invalid codes. 
For example, in Figure~\ref{fig:model2.hoc.stage2}, code at step $\tau=8$ is filtered because the corresponding code commands in Stage 1 terminate on \DSLMove, which is in the middle of the body of loop \DSLRepeatUntil~of $\code^{\textnormal{ref},\ast}$. For the remaining elements, we generate codes from their code commands -- these codes eventually serve as solution codes of the subtasks.
This stage provides a new sequence denoted as $\executiontracefiltered(\task^\textnormal{ref}, \code^{\textnormal{ref},\ast}) := ((\executiontracetask^{\task^\textnormal{ref}}, \code^{\tau}))_{m=1,\ldots,M^\textnormal{filter}}$ where $M^\textnormal{filter}\leq M^\textnormal{all}$. %
Further details of this stage can be found in Appendix~\ref{appendix.model.singlegrid}.

\looseness-1\paragraph{Stage 3: Modifying grids in the trace via symbolic execution (Figure~\ref{fig:model2.hoc.stage3}).}
In this stage, we generate task grids for each of the codes from the sequence obtained in Stage 2. We achieve this using symbolic execution techniques~\citep{DBLP:journals/cacm/King76, DBLP:conf/nips/AhmedCEFGRS20}. Specifically, during symbolic execution, we make minimal modifications to the grids of the subtasks w.r.t. $\task^\textnormal{ref}_{\textnormal{vis},1}$ and generate high quality subtasks. For example, consider step $\tau=6$ in Figures~\ref{fig:model2.hoc.stage2} and \ref{fig:model2.hoc.stage3} . After symbolic execution on the code from Figure~\ref{fig:model2.hoc.stage2}, we obtain the grid in Figure~\ref{fig:model2.hoc.stage3}. Observe how the ``goal'' (red star) is moved to the final location of ``avatar'' on the grid from Figure~\ref{fig:model2.hoc.stage2} to generate the grid in Figure~\ref{fig:model2.hoc.stage3}.
This stage provides a new sequence of subtasks denoted as $\executiontracesymexec(\task^\textnormal{ref}, \code^{\textnormal{ref},\ast}) := ((\task^{\tau}, \code^{\tau}))_{\tau=1,\ldots,M^\textnormal{SE}}$ where $M^\textnormal{SE} \leq M^\textnormal{filter}$. %
Further details of this stage can be found in Appendix~\ref{appendix.model.singlegrid}.

\paragraph{Stage 4: Generating subtasks via subsequence selection.}
\looseness-1In the final stage, we generate a progression of $K'$ subtasks. From \executiontracesymexec, we obtain a set of all subsequences of length $K'$ denoted as $\progressionset(\task^\textnormal{ref},\code^{\textnormal{ref},\ast}, K')$. Using \progressionset, we optimize for Equation~\ref{eq:problemstatement.1} 
to obtain our final progression. Further details of this stage can be found in Appendix~\ref{appendix.model.singlegrid}.
%


\subsection{\synsubtasks}\label{sec.algorithm.alg}

Our synthesis algorithm \synsubtasks~is a combination of procedures described in Sections~\ref{sec.algorithm.multigrid}~and~\ref{sec.algorithm.singlegrid}. Algorithm~\ref{alg:synthesis.short} provides an overview of \synsubtasks~ and we present the implementation details in Appendix~\ref{appendix.sec.model}.
%


\begin{algorithm}[!h]
    \caption{\synsubtasks}
    \begin{algorithmic}[1]
        \Function{\synsubtasks}{$\task^\textnormal{ref}, \code^{\textnormal{ref},\ast}, K$}\label{alg:synthesis.combined}
        \State $\progressionset \leftarrow \{\}$; $\Sigma^{\task^\textnormal{ref}_\textnormal{n}} \leftarrow \text{permutations of }\{1,2,\ldots,\task^\textnormal{ref}_\textnormal{n}\}$
        \For{$\sigma \in \Sigma^{\task^\textnormal{ref}_\textnormal{n}}$}   
        \State $\code^{\textnormal{single},\ast} \leftarrow \reducedcode(\{\task^\textnormal{ref}_{\textnormal{vis},\sigma_1}\}; \task^\textnormal{ref}, \code^{\textnormal{ref},\ast})$ 
        \State $\task^\textnormal{single} := (1, \{\task^\textnormal{ref}_{\textnormal{vis,}\sigma_{1}}\}, \code^{\textnormal{single},\ast}_\textnormal{blocks}, \code^{\textnormal{single},\ast}_\textnormal{size})$
        \State $\progression1 \leftarrow$  $\synsubtasksingle(\task^\textnormal{single}, \code^{\textnormal{single},\ast}, K')$ where $K' = K-\task^\textnormal{ref}_\textnormal{n} + 1$
         \State $\progression2 \leftarrow \synsubtasksmulti(\task^\textnormal{ref}, \code^{\textnormal{ref},\ast}, \sigma)$, i.e., progression for a given permutation $\sigma$ 
            \State $\progression{} \leftarrow$ Concatenate $\progression1$ with $\progression2$ (after removing the common element); add \progression~to \progressionset
        \EndFor{}
        \State \textbf{Return $\progression^{\ast}\in\progressionset$ by optimizing for Equation~\ref{eq:problemstatement.1}}
        \EndFunction
    \end{algorithmic}
    \label{alg:synthesis.short}
\end{algorithm}

%


\section{Experimental Results for Neural Program Synthesis}\label{sec.simulations}

In this section, we first validate the quality of subtasks generated by \synsubtasks~according to the desirable properties of a progression based on metrics introduced in Section~\ref{sec:problemstatement.obj}. Next, we evaluate their utility in assisting neural program synthesizers (NPS) in learning to solve block-based programming tasks. We evaluate the subtasks in the context of NPS agents as a warm-up before evaluating them with novice programmers (discussed subsequently in Section \ref{sec.userstudy}). For the evaluation with NPS agents, we first evaluate the agents trained using supervised learning; then, we further evaluate the performance of agents trained using Reinforcement Learning (RL).

\looseness-1A neural program synthesizer, at a high level, is a neural network-based model that takes a visual task as input, and sequentially synthesizes its solution code~\citep{gulwani2017program, DBLP:conf/iclr/ChenLS19}. We model our NPS agents based on the architecture used in the work of~\citet{DBLP:conf/iclr/BunelHDSK18}. However, these models require large, high-quality, and diverse datasets for training. Inspired by the work of~\citet{DBLP:conf/nips/EysenbachLS21}, we augment the training dataset using our synthesized subtasks to increase its diversity, thereby improving the strength of the learning signals. This also aligns with our objective of evaluating the helpfulness of the subtasks in learning to solve a given reference task. We publicly share the scripts for reproducing the experiments\footref{fn:github-repo}.

\subsection{Methods to Synthesize Progressions and Augmented Datasets}\label{sec.simulations.techniques_and_datasets} 

In this section, we describe the datasets that we use for training NPS agents and explain how we augment them using different subtasking methods. Then, we validate the quality of our subtasking methods by comparing the augmented datasets w.r.t their diversity and quality according to the metrics introduced in Section~\ref{sec:problemstatement.obj}.

\paragraph{Training dataset.} Datasets used in our experiments are based on the work of~\citet{DBLP:conf/iclr/BunelHDSK18}. Specifically, we borrow their training dataset of $\sim{}\!1.1$ million \emph{Karel programming tasks} where each task comprises six visual grids and a reference solution code that solves all six grids. For our experiments, we create smaller datasets of varying sizes by randomly sampling tasks from this dataset. In particular, following \cite{DBLP:conf/iclr/BunelHDSK18}, we create a small dataset of $10{,}000$ tasks. We use this small dataset as the basis for most of our experiments to highlight the benefits of subtask-augmentation when applied to small datasets under limits on the computational resources needed for training. Then, we generate variants of these training datasets by augmenting them with progressions of subtasks synthesized for each training task using four baseline methods. Next, we describe these different subtasking methods.

\paragraph{Methods to synthesize progressions.} 
For NPS training, we compare our subtasking methods, \synsubtasks{} and \synsubtasksmulti{}, with two additional baselines, \nodecomp{} and \harddecomp{}. In our setup, each training task $T^\textnormal{ref}$ has $T^\textnormal{ref}_\textnormal{n} = 6$ and we use $K' = 4$ with $K = 9$. Methods \synsubtasks{} and \synsubtasksmulti{} are described in Section~\ref{sec.algorithm}. \nodecomp{} simply contains $K$ copies of each training task. \harddecomp{} synthesizes a progression of $K$ subtasks such that the solution code of each subtask is the same as that of the reference task, but the visual grids are different. Specifically, \harddecomp{} modifies \synsubtasksingle{} to synthesize subtasks from the execution trace of a given single-grid task such that their solution codes remain the same as that of the task; moreover, it ensures diversity of the visual grids of the subtasks by selecting points uniformly along the execution trace.

\paragraph{Validation on the dataset of $10{,}000$ tasks.} To validate the diversity of the datasets augmented with subtasks generated by different methods, we report the number of unique code complexity values encountered in each augmented set: $104$ for \synsubtasks{}, $89$ for \synsubtasksmulti{}, $78$ for \nodecomp{}, and $89$ for \harddecomp{}. Furthermore, the progressions have the following maximum code complexity jump values (when averaged over all $10{,}000$ tasks), i.e. $\FProgressioncomplexity$ is
$697.7$ for \synsubtasks{}, $743.2$ for \synsubtasksmulti{}, and $945.8$ for \nodecomp{} and \harddecomp{}.
Next, we report the results of training NPS agents with augmented datasets using supervised training approaches and RL-based training approaches.
%


\subsection{Supervised Agents}\label{sec.simulations.supervised}


\begin{figure*}[t!]
\begin{subfigure}[b]{0.65\textwidth}
\begin{minipage}{0.5\textwidth}
\includegraphics[height=4.2cm]{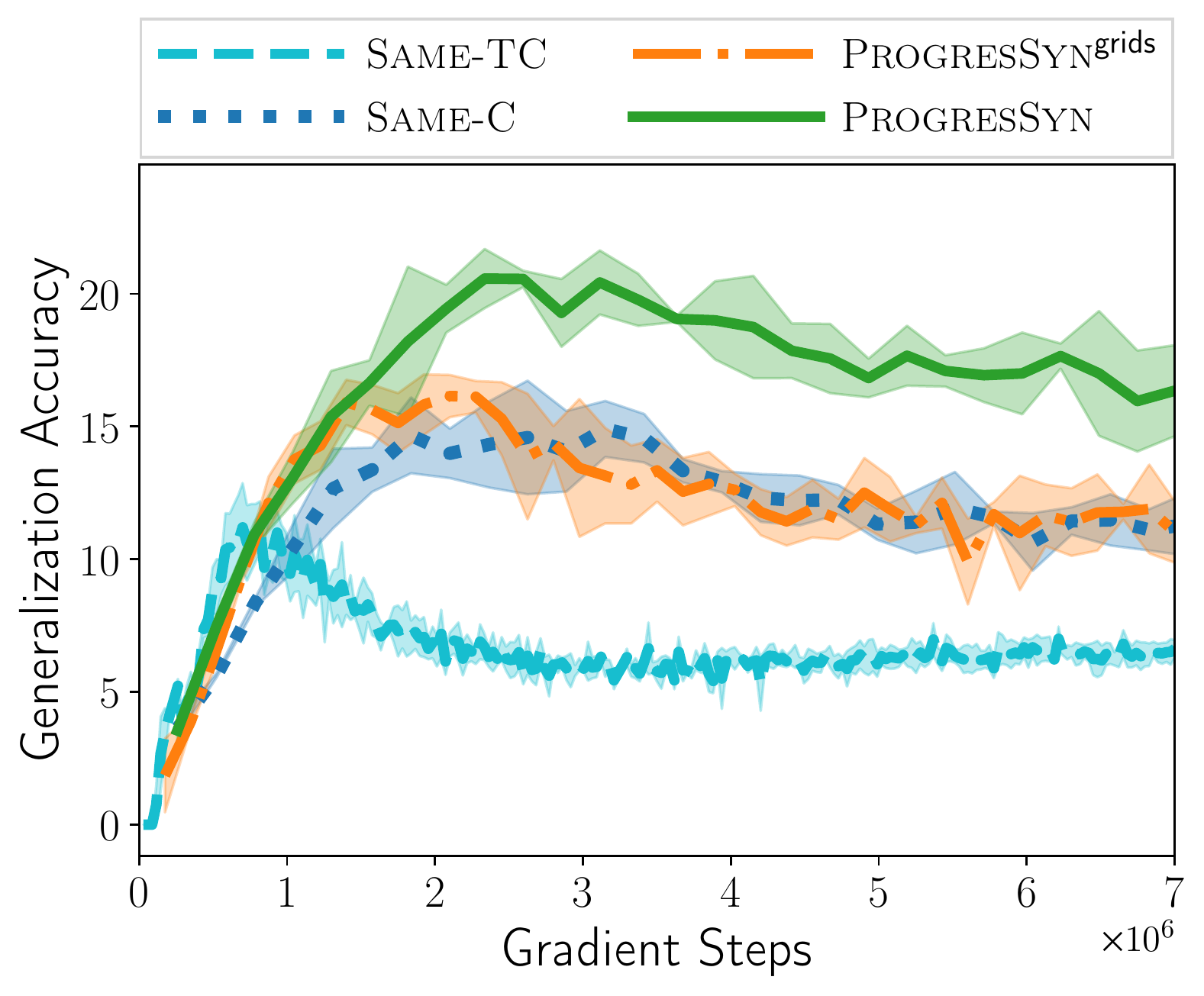}
\end{minipage} 
\ \ \ 
\begin{minipage}{0.5\textwidth}
 \centering
\includegraphics[height=4.2cm]{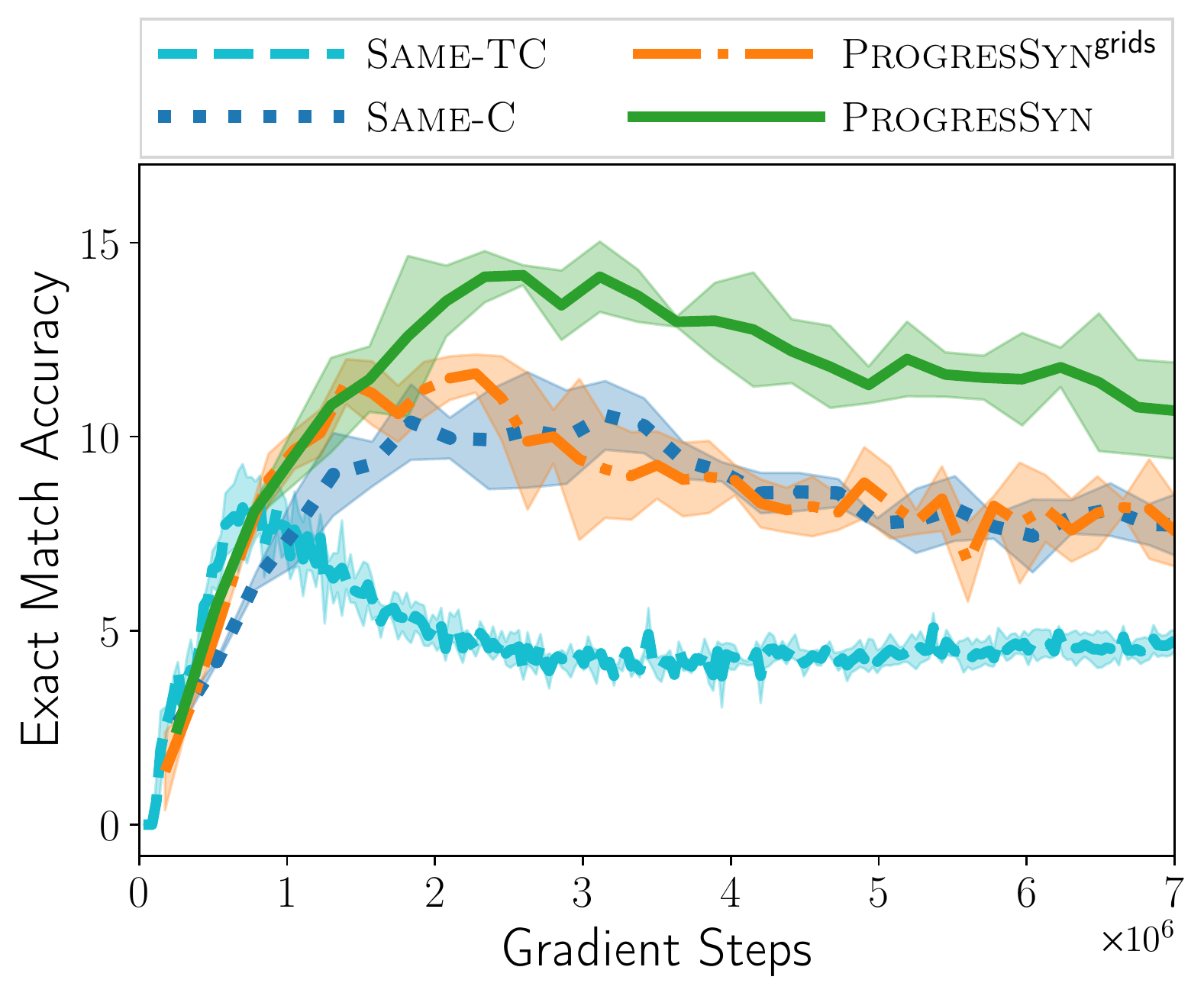}
\end{minipage} 
\caption{Training performance}
\label{results:imitexperiments.plots}
\end{subfigure} 
\ \ \ \ 
\begin{subfigure}[b]{0.33\textwidth}
        \scalebox{0.69}{
	     \renewcommand{\arraystretch}{1.7} 
            \begin{tabular}{r||rrrr}
        			\toprule
        			Method & 
        			Exact &
        			Generalization 
        			 \\
                       [0.25cm]
        			\toprule
        			\defaultTraining & $8.7 \pm 0.8$ & $11.8 \pm 1.3$
        			\\
                     [0.25cm] {\textsc{Same-}\code} & $10.8 \pm 0.6$ & $15.0 \pm 0.8$
                        \\
                    [0.25cm]
        	\ours\textsuperscript{grids} & $12.1 \pm 1.0$ & $16.8 \pm 1.4$ 
                    \\
                    [0.25cm]
                    \ours & $15.2 \pm 0.4$ & $21.6 \pm 0.5$
        			\\
        			\bottomrule
          \end{tabular}
          }
\vspace{5.5mm}
\caption{Final performance}
\label{results:imitexperiments.final}       
\end{subfigure}
\caption{This figure shows the performance of a neural program synthesizer based on~\cite{DBLP:conf/iclr/BunelHDSK18} and trained in a supervised-fashion by augmenting the pruned dataset of $10{,}000$ tasks with subtasks generated using different methods. Results are reported as mean and standard deviation over $3$ random seeds. Plots in \textbf{(a)} show the generalization and exact match accuracy during training when evaluated on the validation dataset of $2500$ tasks from \cite{DBLP:conf/iclr/BunelHDSK18}; the x-axis shows the number of gradient steps. Table in \textbf{(b)} shows the final performance of the model obtained for each method when evaluated on the test dataset of $2500$ tasks from \cite{DBLP:conf/iclr/BunelHDSK18}. }
\label{results:imitexperiments}
\end{figure*}

%


\begin{figure*}[t!]
\begin{subfigure}[b]{1.\textwidth}
\begin{minipage}{0.5\textwidth}
\centering
\includegraphics[height=4.6cm]{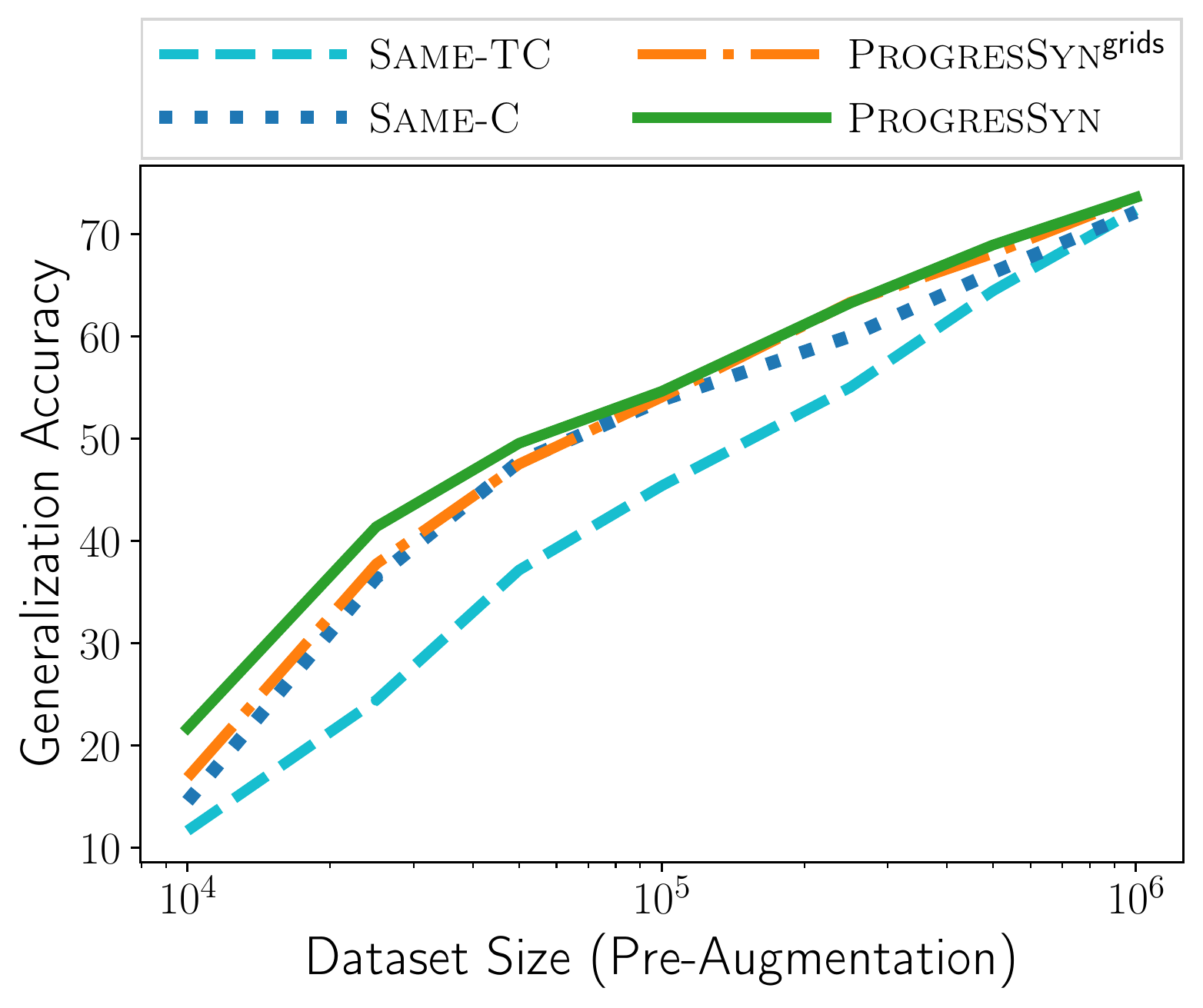}
\end{minipage} 
\ \ \ 
\begin{minipage}{0.5\textwidth}
 \centering
\includegraphics[height=4.6cm]{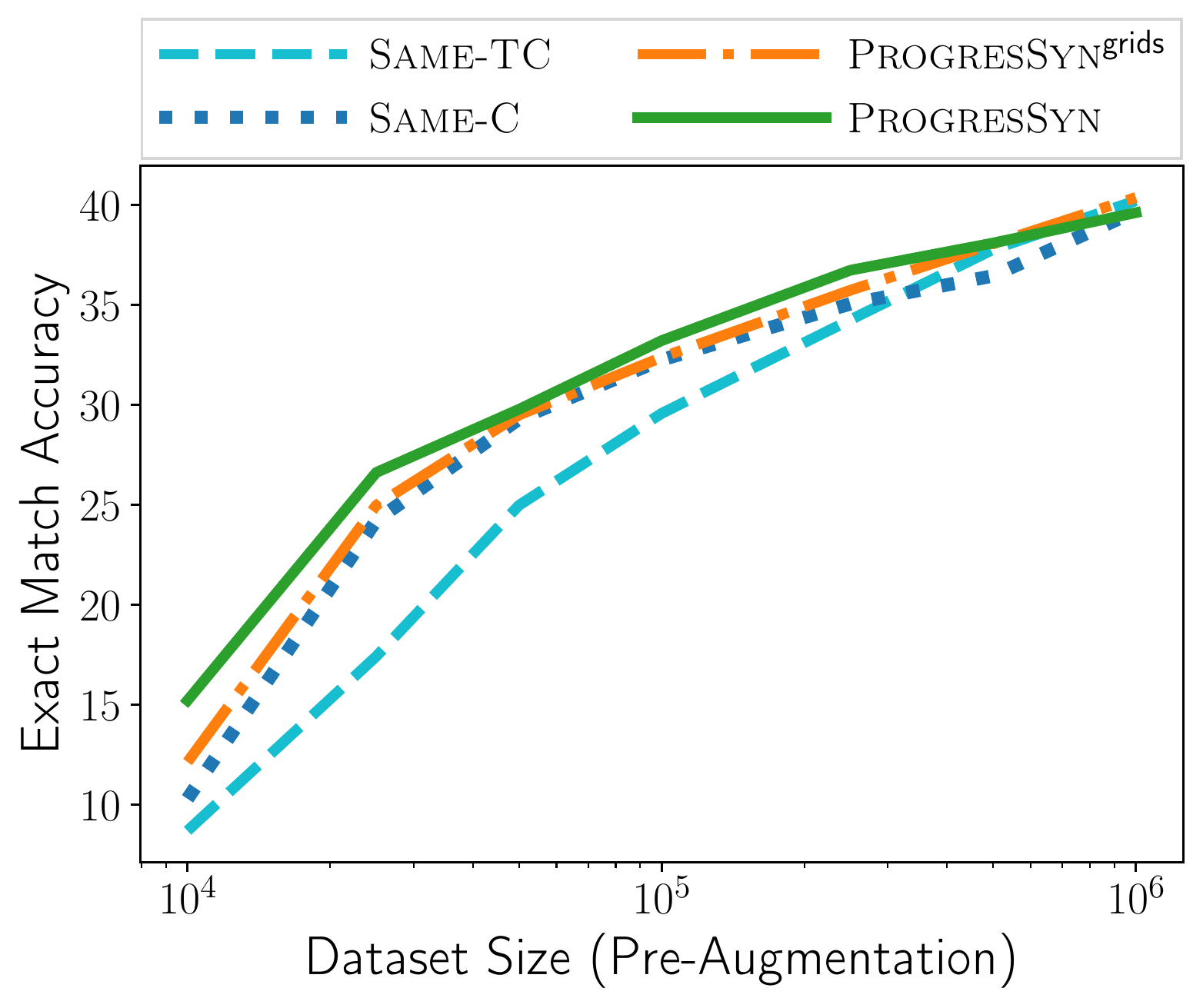}
\end{minipage} 
\label{results:datasetsize_experiments.plots}
\end{subfigure} 
%
\caption{This figure demonstrates the performance of neural program synthesizers after supervised training using datasets of varying sizes augmented by methods described in Section~\ref{sec.simulations.techniques_and_datasets}. Plots show the generalization and exact match accuracy of the best models obtained for each method when evaluated on the test dataset of 2500 tasks from \cite{DBLP:conf/iclr/BunelHDSK18}. The x-axis shows the size of the training dataset before augmentation with the subtasks generated using different methods.}
\label{results:datasetsize_experiments}
\end{figure*}
%

In this section, we show that augmenting training datasets with progressions generated by our subtask synthesis methods improve the performance of NPS agents. Specifically, we investigate the performance gain from using subtask-augmented training datasets during supervised training of NPS agents. We begin with supervised training, which is computationally less expensive and allows easier experimentation with larger dataset sizes.

\paragraph{Setup.}  Our experimental setup is based on the work of~\citet{DBLP:conf/iclr/BunelHDSK18}. We use their Maximum Likelihood Estimated (MLE)-based supervised training component for training the NPS agents. Our neural program synthesizer is also modeled based on the neural architecture from the work of~\citet{DBLP:conf/iclr/BunelHDSK18}. 
For the syntax model of the codes, we used their "learned syntax module" approach as it performs better than the "rule-based syntax module" approach when evaluated for datasets of varying sizes. While we borrow the experimental setup from the work of~\citet{DBLP:conf/iclr/BunelHDSK18}, our setup differs from them in one significant way: We pruned tasks from the training dataset of $10{,}000$ tasks where the solution code did not have full coverage w.r.t. the task (e.g., tasks whose solution code had redundant blocks). This was done because our approach requires access to good-quality reference tasks whose solution codes have full coverage. Hence, the final training dataset  comprised $7{,}300$ tasks. We trained our models for $100$ million gradient steps.\footnote{Training was done on a single 32GB Tesla V100S GPU and took approximately $8$ days.}

\paragraph{Results.} We first report the results of training NPS agents using augmented variants of the dataset of $10{,}000$ tasks in Figure~\ref{results:imitexperiments}, averaged over $3$ random seeds. The plots show that agents trained with a more diverse training dataset saturate later and reach higher peak performance. The final performance of agents trained with \synsubtasks{} is over $80\%$ higher than those trained with \nodecomp. The final performance of \synsubtasks{} is also higher than that of \synsubtasksmulti{} highlighting the utility of more fine-grained subtasking. \synsubtasks{} also outperforms \harddecomp{}, indicating the importance of well-spaced code complexity in the progression.
Furthermore, our results also generalize to larger datasets (as indicated by the final performance of the synthesizers trained on augmented datasets of different sizes in Figure~\ref{results:datasetsize_experiments}). Figure~\ref{results:datasetsize_experiments} shows that while performance gains from augmenting the dataset diminish when the original dataset is very large before augmentation, the benefits observed from training with smaller datasets still generalize to the setting where agents are trained with larger datasets. For example, the performance gain w.r.t generalization accuracy from training with datasets augmented with progressions synthesized by \synsubtasks{} compared to \nodecomp{} is $70\%$ for $25,000$ tasks, $33\%$ for $50,000$ tasks, $20\%$ for $100,000$ tasks, $15\%$ for $250,000$ tasks, and $7\%$ for $500,000$ tasks.

%


%
\subsection{RL Agents}\label{sec.simulations.rl} 

In this section we evaluate the performance of our subtasking method when NPS agents are trained using reinforcement learning (henceforth referred to as RL agents). \citet{DBLP:conf/iclr/BunelHDSK18} shows that RL leads to performance gains over supervised training of NPS models, especially when applied to small datasets.
In this work, we also focus on small datasets when evaluating the performance of our subtasking method with RL agents.

\begin{figure*}[t!]
\begin{subfigure}[b]{0.65\textwidth}
\begin{minipage}{0.5\textwidth}
            \includegraphics[height=4.2cm]{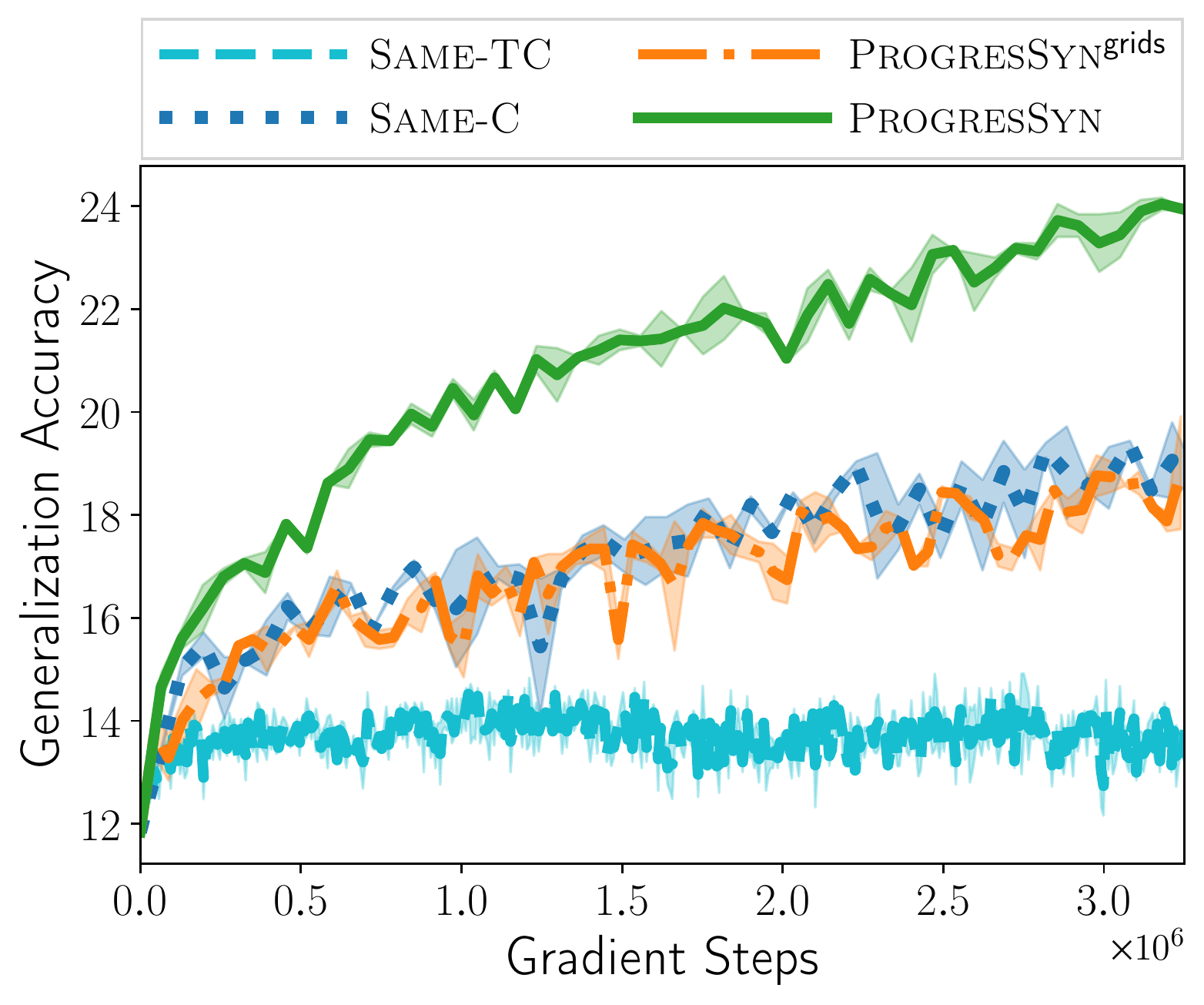}
\end{minipage} 
\ \ \ 
\begin{minipage}{0.5\textwidth}
 \centering
            \includegraphics[height=4.2cm]{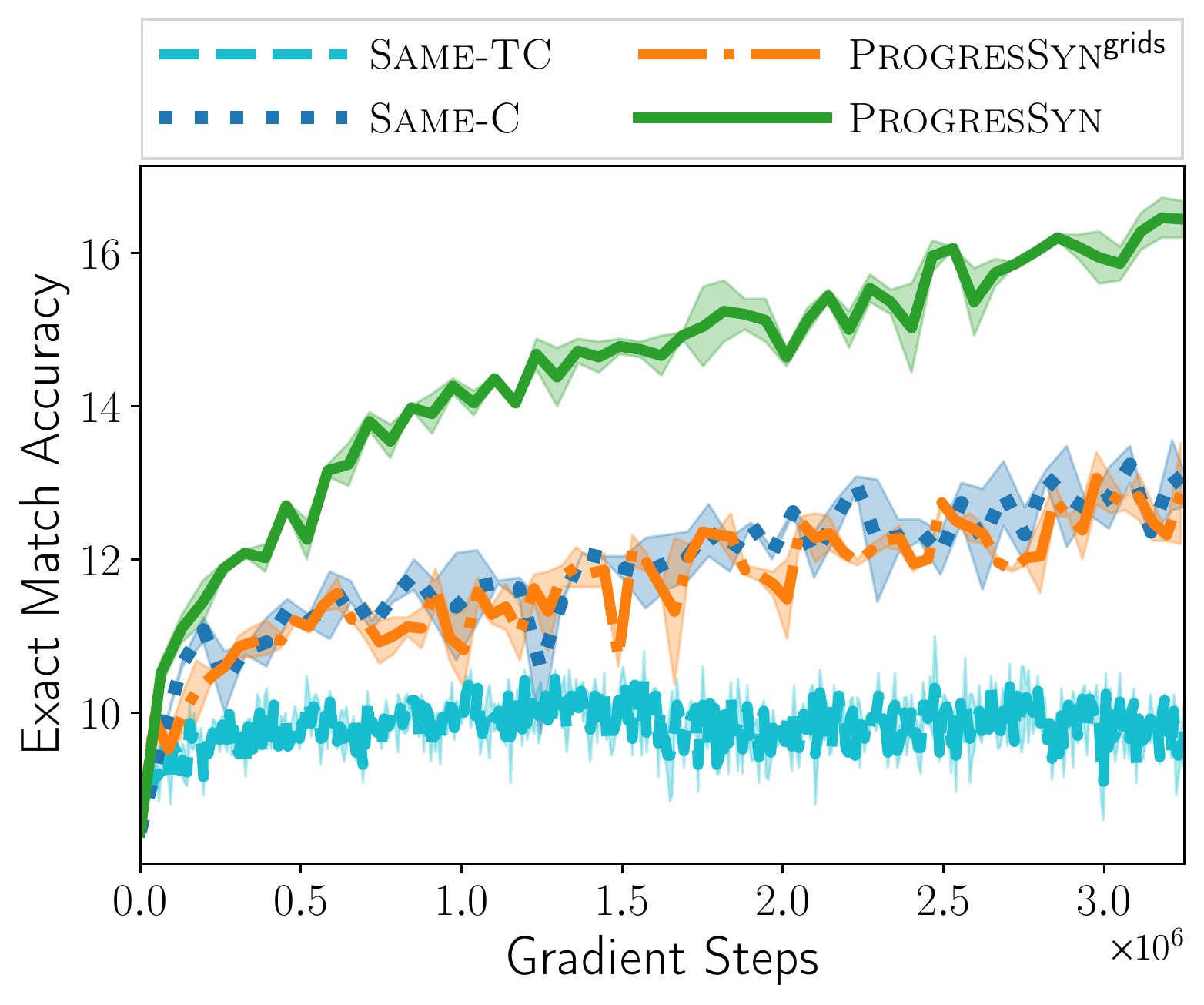}
\end{minipage} 
\caption{Training performance}
\label{results:rlexperiments.plots}
\end{subfigure} 
\ \ \ \ 
\begin{subfigure}[b]{0.33\textwidth}
        \scalebox{0.69}{
	     \renewcommand{\arraystretch}{1.7} 
            \begin{tabular}{r||rrrr}
        			\toprule
        			Method & 
        			Exact &
        			Generalization 
        			 \\
                       [0.25cm]
        			\toprule
        			\defaultTraining & $10.6 \pm 0.1$ & $14.7 \pm 0.2$
        			\\
                     [0.25cm] {\textsc{Same-}\code} & $13.7 \pm 0.2$ & $19.2 \pm 0.3$
                        \\
                    [0.25cm]
        	\ours\textsuperscript{grids} & $13.3 \pm 0.3$ & $18.8 \pm 0.5$ 
                    \\
                    [0.25cm]
                    \ours & $17.3 \pm 0.3$ & $24.6 \pm 0.5$
        			\\
        			\bottomrule
          \end{tabular}
          }
\vspace{5.5mm}
\caption{Final performance}
\label{results:rlexperiments.final}       
\end{subfigure}
\caption{This figure parallels Figure~\ref{results:imitexperiments} but the synthesizers are trained using RL instead of supervised learning. Plots in \textbf{(a)} show the generalization and exact match accuracy during training when evaluated on the validation dataset of $2500$ tasks from \cite{DBLP:conf/iclr/BunelHDSK18}; the x-axis shows the number of gradient steps. Table in \textbf{(b)} shows the final performance of the model obtained for each method when evaluated on the test dataset of $2500$ tasks from \cite{DBLP:conf/iclr/BunelHDSK18}.}
\label{results:rlexperiments}
\end{figure*}

%

%

\paragraph{Setup.} Our experimental setup with RL agents is the same as the setup described in Section~\ref{sec.simulations.supervised} except that we adopt the RL-based training approach of~\citet{DBLP:conf/iclr/BunelHDSK18}. Also, we trained our models for $3.25$ million gradient steps instead of $100$ million and limited our training dataset to the smaller dataset of $10{,}000$ tasks due to the computational complexity of RL training.

\textbf{Results.}
We report the results of training RL agents using augmented datasets in Figure~\ref{results:rlexperiments}, averaged over $3$ random seeds. The plots show that agents trained with a more diverse training dataset have a steeper learning curve. The final performance of agents trained with \synsubtasks{} is over $60\%$ higher than those trained with \nodecomp. The final performance of \synsubtasks{} is also higher than that of \synsubtasksmulti{} and \harddecomp{}, indicating the importance of more fine-grained subtasking while synthesizing the progression. These results are analogous to our results obtained from training supervised agents. (see Section~\ref{sec.simulations.supervised}).

\paragraph{Limitations and possible extensions.} Our results show that datasets augmented with our synthesis methods significantly benefit when used with RL training. This is particularly promising as RL opens up many new possibilities to incorporate the subtasks in the training process. For example, they could be used to explicitly design intermediate rewards which could yield similar performance gains with lower training time. Furthermore, in our current setup, we do not explicitly consider the sequence of the progression, i.e. we do not exploit the well-spaced code complexity of subtasks in the progression. However, the progression provides a natural curriculum for training the RL agents. Using such a curriculum could further improve their training time and performance.

%


\section{Assisting Novice Human Programmers}\label{sec.userstudy}

In this section, we evaluate the utility of the progression of subtasks synthesized by \ours{} in assisting novice programmers in solving block-based visual programming tasks taken from real-world programming platforms. We first describe the setup used for conducting the user study. Specifically, we developed an online platform for the study. Next, we present the different methods evaluated on the platform. We also present three research questions to analyze the desirable properties of our progression. Finally, we present the results of our study centered around our research questions.

\subsection{Setup}\label{sec.userstudy.setup}
\looseness-1\paragraph{Participants.} We recruited participants for the study from Amazon Mechanical Turk; an IRB approval had been obtained for the study. The participants were US-based adults, without expertise in block-based visual programming. The study took at most 30-35 minutes to complete, per participant. Due to the costs involved (over $4$ USD per task for a participant), we used only two reference tasks for the study. These tasks are based on real-world tasks from \emph{Hour of Code: Maze Challenge} by \emph{Code.org}~\citep{hourofcode_maze}. Specifically, we used the following tasks: Maze$08$ ($\FTaskcomplexity = 1005$) and Maze$16$ ($\FTaskcomplexity = 2004$). Henceforth, we refer to these tasks as \hocd{} and \hocg{} respectively. \hocd{} is illustrated in Figure~\ref{fig:appendix.hoc.og} and \hocg{} is illustrated in Figure~\ref{fig:intro.hoc.og}.

\paragraph{Online platform.} We developed a web app for the study (see link in Footnote~\ref{fn:github-repo}). On the app we have enabled our subtasking algorithm \ours{} and three block-based visual programming tasks: \hocd{}, \hocg{} and \emph{Karel programming environment} based \textsc{Stairway} (see Figure~\ref{fig:intro.multigrid.og}). Note that, while the app contains three reference tasks, the user study was conducted only with reference tasks \hocd{} and \hocg{}. The app uses the publicly available toolkit of Blockly Games~\citep{blocklygames} and provides an interface for a participant to solve a block-based visual programming task through a progression of subtasks. Before logging into the app, each participant was encouraged to watch a $4$ minute instructional video about block-based programming to familiarize themselves with the platform. After logging into the app, a participant was assigned a reference task $\task^\textnormal{ref} \in \{\hocd{}, \hocg{}\}$ and one of the subtasking methods 
at random. These elements constituted a ``session'' for a participant. Specifically, a participation session comprised of the following steps: (i) Step $1$: The participant is shown the reference task, and given $10$ attempts to solve it. If they are successful, they exit the platform; otherwise, they proceed to the next step; (ii) Steps $2a$ / $2b$: The participant is presented with the first $K-1$ (i.e., $3-1=2$) subtasks from the progression synthesized by the assigned subtasking method, and given $10$ attempts to solve each subtask; (iii) Step $3$: The participant is presented with the reference task from Step $1$ again, and given $10$ attempts to solve it. 
Each of these steps are illustrated in Appendix~\ref{appendix.userstudy.app.details}.

\subsection{Methods evaluated and Research Questions}\label{sec.userstudy.rqs}
Similar to our evaluation with neural program agents (described in Section~\ref{sec.simulations}), we evaluate the performance of \synsubtasks{} in comparison to the baseline methods, \default{}, \nodecomp{}, \harddecomp{} and \human{}. \human{} is the additional baseline we use for the user study. For both our reference tasks, $\task^\textnormal{ref}_\textnormal{n}=1$; we use $K'=3$ and $K=3$. We describe these methods below.

\paragraph{Methods evaluated.} Method \default{} is the setting where the participant is presented with $\task^\textnormal{ref}$ and given $10$ tries to solve it (only Step $1$ of the participation session). Methods \nodecomp{} and \harddecomp{} (collectively called \noharddecomp{}) generate a progression of $3$ subtasks which are not well spaced w.r.t their code complexity. Specifically, \nodecomp{} contains $3$ subtasks all of which are the same as the reference task. \harddecomp{} minimally alters the visual grids of the subtasks while their solution codes remain the same as that of the reference task. Methods \expnew{} and \expopt{} (collectively called \human{}) generate a handcrafted progression of $3$ subtasks, that are well spaced w.r.t. their code complexity but have task grids that do not retain the visual context of the reference task. Figure~\ref{fig:userstudy.hoc} illustrates these methods on reference task \hocg{} and Figure~\ref{fig:appendix.userstudy.hoc} illustrates these methods on reference task \hocd{}. We note that a participant spends up to $40$ problem-solving attempts in all the subtasking methods (\noharddecomp{}, \human{}, \ours{}) and up to $10$ problem-solving attempts without subtasking (\default{}). We also present additional details about progressions synthesized by these methods w.r.t the desirable properties of subtasks (as described in Section~\ref{sec:problemstatement.obj}) in Appendix~\ref{appendix.userstudy_validation}. Next, we present the research questions around which our study was designed. 

\paragraph{Research questions.} We center our user study around the following research questions \emph{(RQs)} to measure the efficacy of \ours{}: (i) \textit{RQ1: Usefulness of subtasking.} Does solving a progression of subtasks increase the success rate on the reference task? (ii) \textit{RQ2: Well-spaced code complexity.} Do progressions with subtasks that are well spaced w.r.t. their code complexity improve the success rate more in comparison to progressions that violate this property?
(iii) \textit{RQ3: Retaining visual context of the reference task.} Do progression of subtasks that retain the visual context of the reference task in their grids improve the success rate more in comparison to progressions that violate this property?



\begin{figure*}[t!]
\centering
    \scalebox{0.9}{
    \renewcommand{\arraystretch}{1.2}
    \begin{tabular}{r||rrr||rrr}
			\toprule
			Method & \multicolumn{3}{c||}{Total participants} & \multicolumn{3}{c}{Fraction succeeded} 
			\\
			 & All  & \hocd & \hocg &
			 All  & \hocd & \hocg
			 \\
			\toprule
			\default & $114$ & $57$ & $57$ 
			& $0.605$ & $0.807$ & $0.403$ 
			\\
 			\hline
			\nodecomp & $114$ & $57$ & $57$ 
                & $0.667$ & $0.842$ & $0.491$ 
                \\
                \harddecomp & $116$ & $59$ & $57$ 
                & $0.672$ & $0.847$ & $0.491$ 
                \\
    			\noharddecomp  & $230$ & $116$ & $114$ 
                & $0.669$ & $0.845$ & $0.491$ \\
            \hline
            \human & $235$ & $117$ & $118$ 
            & $0.647$ & $0.838$ & $0.458$ 
            \\ \hline
			\ours & $117$ & $58$ & $59$ 
			& $0.701$ & $0.862$ & $0.542$ 
            \\
			\bottomrule
  \end{tabular}
  }
\caption{Results on tasks \emph{Maze: 08} (referred to as \hocd) and \emph{Maze: 16} (referred to as \hocg) taken from the \emph{Hour of Code: Maze Challenge}. ``All'' represents aggregated results.}
\label{table:userstudy}
\end{figure*}



\begin{figure*}[!t]
\centering
     \begin{subfigure}[b]{0.32\textwidth}
    \centering
    {
        \includegraphics[height=2.95cm]{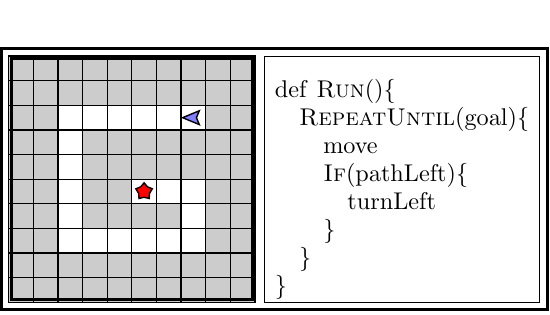}
        \caption{Reference task: $(\task^\textnormal{ref}, \code^{\textnormal{ref},\ast})$}
        \label{fig:sec.userstudy.hocg}
    }
    \end{subfigure}
    \vspace{-1.7mm}
    \\
    \begin{subfigure}[b]{0.32\textwidth}
    \centering
    {
        \includegraphics[height=2.95cm]{./figs/userstudy/userstudy_hoc_nodecomp1.pdf}
        \caption{\nodecomp: $(\task^{1}, \code^{1,\ast})$}
    }
    \end{subfigure}
    \begin{subfigure}[b]{0.32\textwidth}
        \centering
        {
            \includegraphics[height=2.95cm]{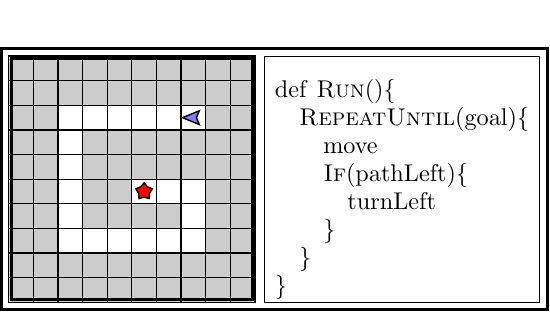}
            \caption{\textbf{}\nodecomp: $(\task^{2}, \code^{2,\ast})$}
        }
    \end{subfigure}
    \begin{subfigure}[b]{0.32\textwidth}
        \centering
        {
            \includegraphics[height=2.95cm]{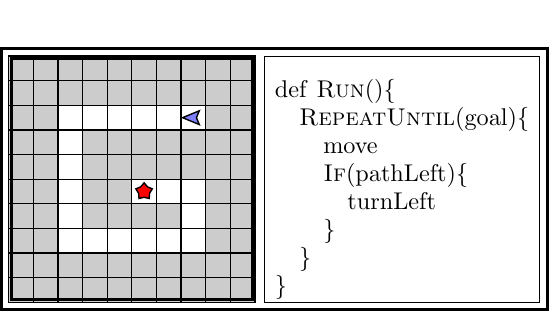}
            \caption{\nodecomp: $(\task^{3}, \code^{3,\ast})$}
        }
    \end{subfigure}
    \vspace{-1.7mm}
    \\
    \begin{subfigure}[b]{0.32\textwidth}
    \centering
    {
        \includegraphics[height=2.95cm]{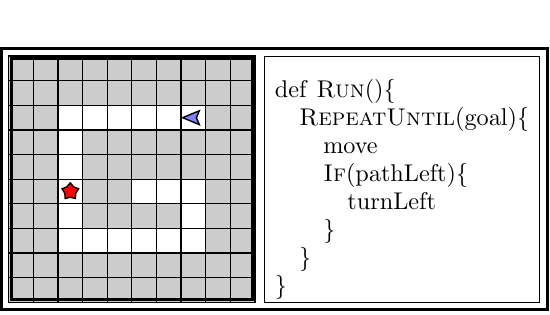}
        \caption{\harddecomp: $(\task^{1}, \code^{1,\ast})$}
    }
    \end{subfigure}
    \begin{subfigure}[b]{0.32\textwidth}
        \centering
        {
            \includegraphics[height=2.95cm]{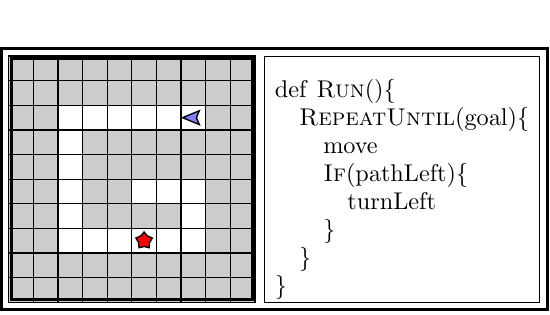}
            \caption{\harddecomp: $(\task^{2}, \code^{2,\ast})$}
        }
    \end{subfigure}
    \begin{subfigure}[b]{0.32\textwidth}
        \centering
        {
            \includegraphics[height=2.95cm]{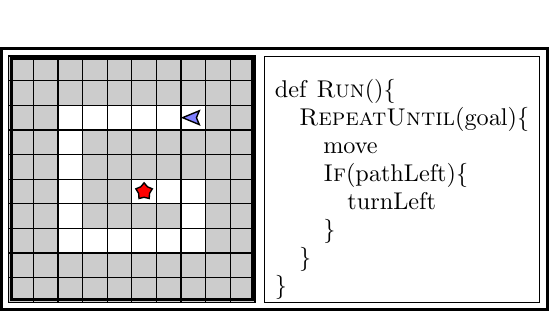}
            \caption{\harddecomp: $(\task^{3}, \code^{3,\ast})$}
        }
    \end{subfigure}
    \vspace{-1.7mm}
    \\
    \begin{subfigure}[b]{0.32\textwidth}
        \centering
        {
            \includegraphics[height=2.95cm]{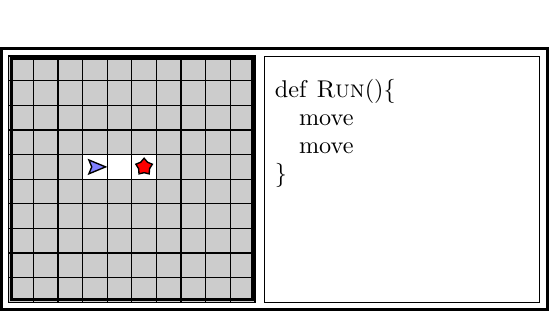}
            \caption{\expnew: $(\task^{1}, \code^{1,\ast})$}
        }
    \end{subfigure}
    \begin{subfigure}[b]{0.32\textwidth}
        \centering
        {
            \includegraphics[height=2.95cm]{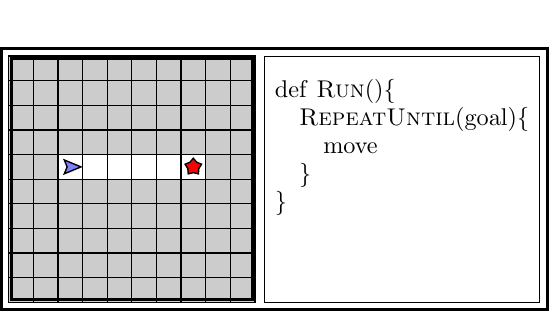}
            \caption{\expnew: $(\task^{2}, \code^{2,\ast})$}
        }
    \end{subfigure}
    \begin{subfigure}[b]{0.32\textwidth}
        \centering
        {
            \includegraphics[height=2.95cm]{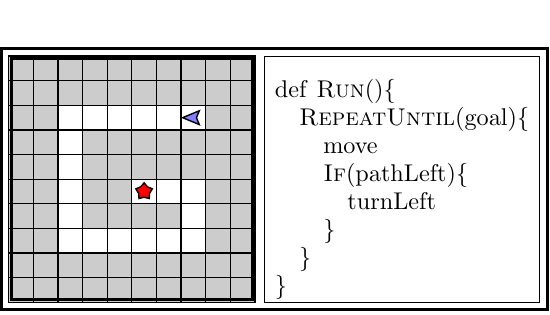}
            \caption{\expnew: $(\task^{3}, \code^{3,\ast})$}
        }
    \end{subfigure}
    \vspace{-1.7mm}
    \\
    \begin{subfigure}[b]{0.32\textwidth}
        \centering
        {
            \includegraphics[height=2.95cm]{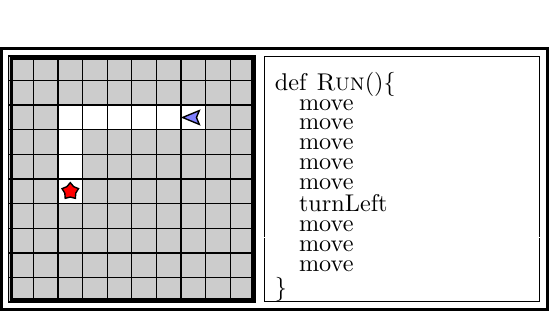}
            \caption{\expopt: $(\task^{1}, \code^{1,\ast})$}
        }
    \end{subfigure}
    \begin{subfigure}[b]{0.32\textwidth}
        \centering
        {
            \includegraphics[height=2.95cm]{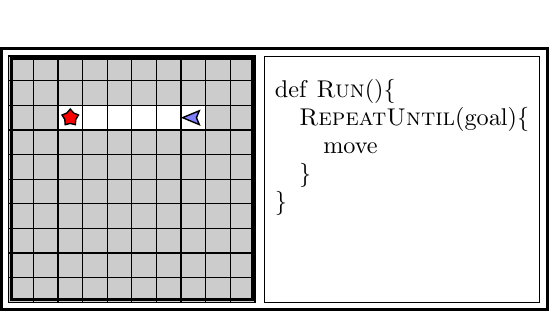}
            \caption{\expopt: $(\task^{2}, \code^{2,\ast})$}
        }
    \end{subfigure}
    \begin{subfigure}[b]{0.32\textwidth}
        \centering
        {
            \includegraphics[height=2.95cm]{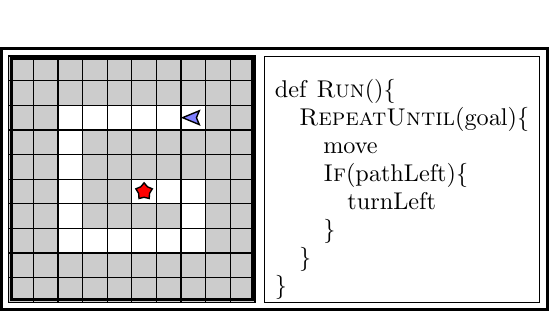}
            \caption{\expopt: $(\task^{3}, \code^{3,\ast})$}
        }
    \end{subfigure}
    \vspace{-1.7mm}
    \\
    \begin{subfigure}[b]{0.32\textwidth}
        \centering
        {
            \includegraphics[height=2.95cm]{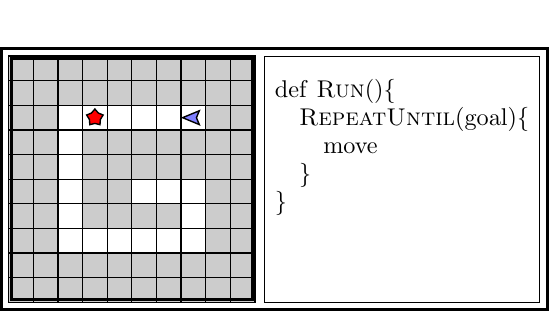}
            \caption{\ours: $(\task^{1}, \code^{1,\ast})$}
        }
    \end{subfigure}
    \begin{subfigure}[b]{0.32\textwidth}
        \centering
        {
            \includegraphics[height=2.95cm]{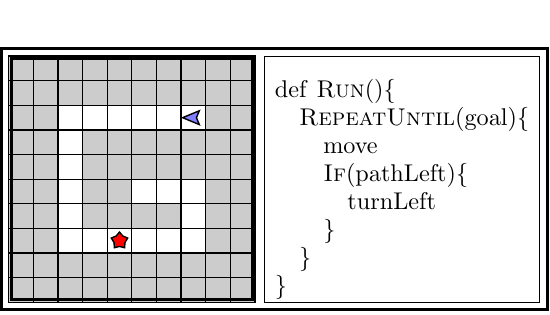}
            \caption{\ours: $(\task^{2}, \code^{2,\ast})$}
        }
    \end{subfigure}
    \begin{subfigure}[b]{0.32\textwidth}
        \centering
        {
            \includegraphics[height=2.95cm]{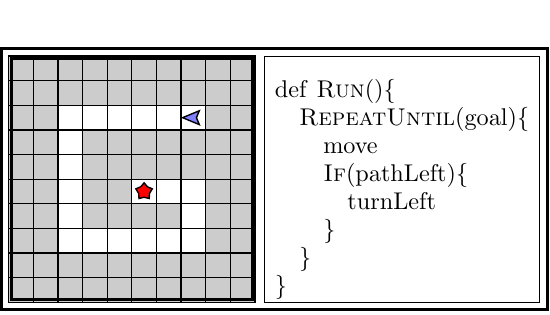}
            \caption{\ours: $(\task^{3}, \code^{3,\ast})$}
        }
    \end{subfigure}
    \caption{Illustration of subtasks synthesized by baseline methods \nodecomp{}, \harddecomp{}, \expnew{}, \expopt{}, and \ours{} for the reference task \hocg{} shown in (a). Each method synthesizes a progression of three subtasks.
    \nodecomp~and \harddecomp~are collectively called \noharddecomp; \expnew~and \expopt~are collectively called \human{}. See \iftoggle{SuppContentOnly}{Section 5 and  Appendix~\ref{appendix.sec.userstudy}}{Section~\ref{sec.userstudy} and Appendix~\ref{appendix.sec.userstudy}} for details.
    }
    \label{fig:userstudy.hoc}
    \vspace{-2.5mm}
\end{figure*}

%


\begin{figure*}[!t]
\centering
    \begin{subfigure}[b]{1\textwidth}
        \centering
        {
            \includegraphics[height=3.05cm]{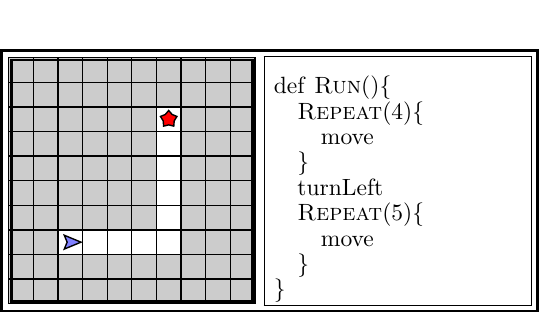}
            \vspace{-1.2mm} 
            \caption{Reference task: $(\task^\textnormal{ref}, \code^{\textnormal{ref},\ast})$}
            \label{fig:appendix.hoc.og}
            \vspace{-2.9mm}
        }
    \end{subfigure}
    \\
    \begin{subfigure}[b]{0.32\textwidth}
        \centering
        {
            \includegraphics[height=3.05cm]{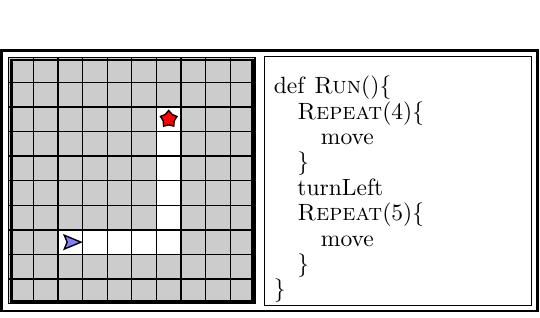}
            \vspace{-4mm}            
            \caption{\nodecomp: $(\task^{1},\code^{1,\ast})$}
            \vspace{-2.9mm}
        }
    \end{subfigure}
    \begin{subfigure}[b]{0.32\textwidth}
        \centering
        {
        \includegraphics[height=3.05cm]{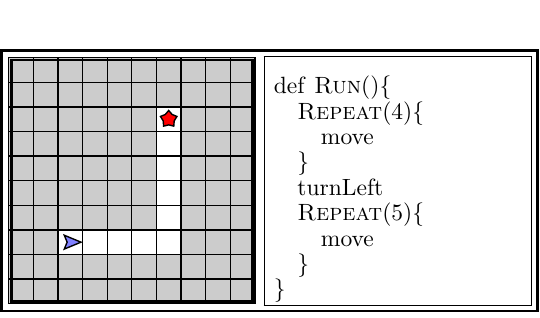}
        \vspace{-4mm}        
        \caption{\nodecomp: $(\task^{2}, \code^{2,\ast})$}
        \vspace{-2.9mm}
    }
    \end{subfigure}
    \begin{subfigure}[b]{0.32\textwidth}
        \centering
        {
            \includegraphics[height=3.05cm]{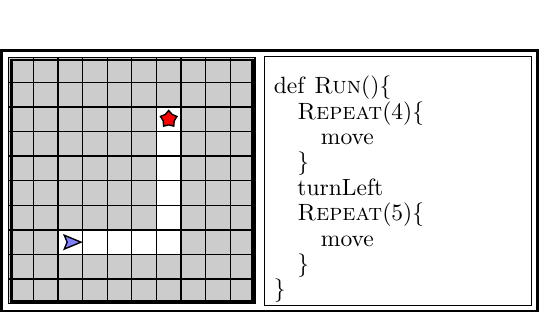}
            \vspace{-4mm}            
            \caption{\nodecomp: $(\task^{3}, \code^{3,\ast})$}
            \vspace{-2.9mm}
        }
    \end{subfigure}
    \\
    \begin{subfigure}[b]{0.32\textwidth}
    \centering
    {
        \includegraphics[height=3.05cm]{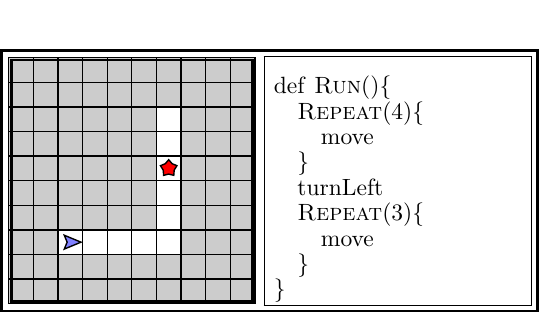}
        \vspace{-4mm}        
        \caption{\harddecomp: $(\task^{1}, \code^{1,\ast})$}
        \vspace{-2.9mm}
    }
    \end{subfigure}
    \begin{subfigure}[b]{0.32\textwidth}
        \centering
        {
            \includegraphics[height=3.05cm]{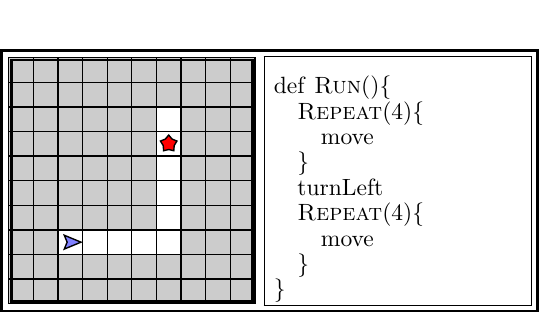}
            \vspace{-4mm}            
            \caption{\harddecomp: $(\task^{2}, \code^{2,\ast})$}
            \vspace{-2.9mm}
        }
    \end{subfigure}
    \begin{subfigure}[b]{0.32\textwidth}
        \centering
        {
            \includegraphics[height=3.05cm]{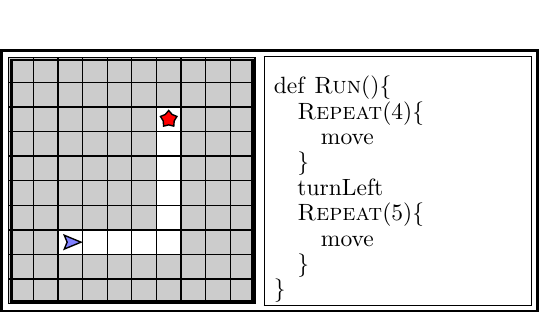}
            \vspace{-4mm}            
            \caption{\harddecomp: $(\task^{3}, \code^{3,\ast})$}
            \vspace{-2.9mm}
        }
    \end{subfigure}
    \\
    \begin{subfigure}[b]{0.32\textwidth}
        \centering
        {
            \includegraphics[height=3.05cm]{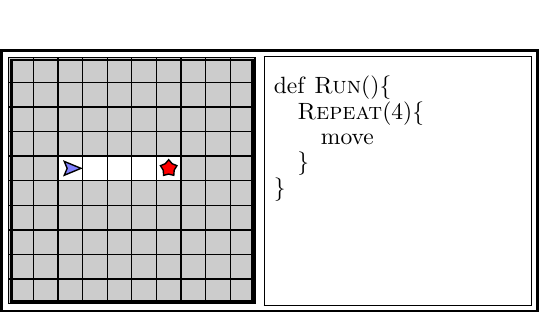}
            \vspace{-4mm}            
            \caption{\expnew: $(\task^{1}, \code^{1,\ast})$}
            \vspace{-2.9mm}
        }
    \end{subfigure}
    \begin{subfigure}[b]{0.32\textwidth}
        \centering
        {
            \includegraphics[height=3.05cm]{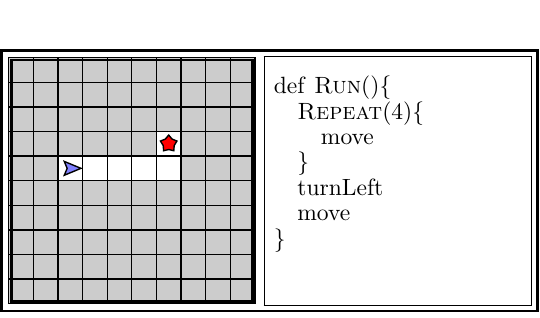}
            \vspace{-4mm}            
            \caption{\expnew: $(\task^{2}, \code^{2,\ast})$}
            \vspace{-2.9mm}
        }
    \end{subfigure}
    \begin{subfigure}[b]{0.32\textwidth}
        \centering
        {
            \includegraphics[height=3.05cm]{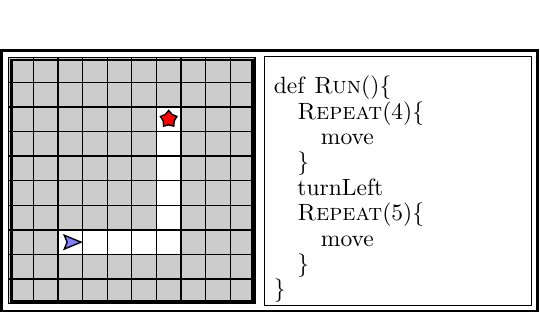}
            \vspace{-4mm}
            \caption{\expnew: $(\task^{3}, \code^{3,\ast})$}
            \vspace{-2.9mm}
        }
    \end{subfigure}
    \\
    \begin{subfigure}[b]{0.32\textwidth}
        \centering
        {
            \includegraphics[height=3.05cm]{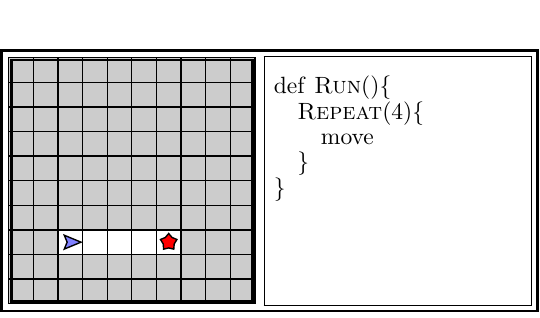}
            \vspace{-4mm}
            \caption{\expopt: $(\task^{1}, \code^{1,\ast})$}
            \vspace{-2.9mm}
        }
    \end{subfigure}
    \begin{subfigure}[b]{0.32\textwidth}
        \centering
        {
            \includegraphics[height=3.05cm]{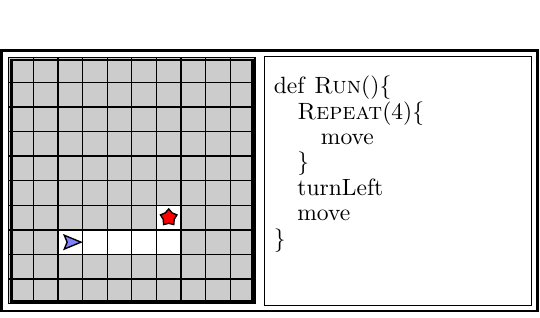}
            \vspace{-4mm}
            \caption{\expopt: $(\task^{2}, \code^{2,\ast})$}
            \vspace{-2.9mm}
        }
    \end{subfigure}
    \begin{subfigure}[b]{0.32\textwidth}
        \centering
        {
            \includegraphics[height=3.05cm]{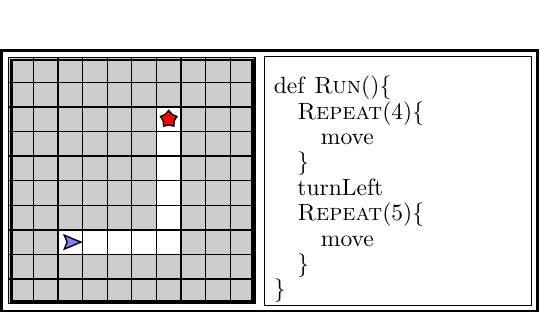}
            \vspace{-4mm}
            \caption{\expopt: $(\task^{3}, \code^{3,\ast})$}
            \vspace{-2.9mm}
        }
    \end{subfigure}
    \\
    \begin{subfigure}[b]{0.32\textwidth}
        \centering
        {
            \includegraphics[height=3.05cm]{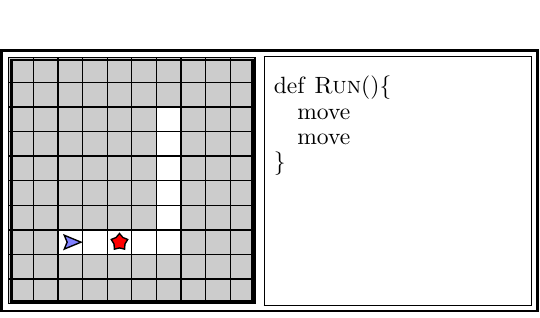}
            \vspace{-4mm}
            \caption{\ours: $(\task^{1}, \code^{1,\ast})$}
            \vspace{-2.9mm}
        }
    \end{subfigure}
    \begin{subfigure}[b]{0.32\textwidth}
        \centering
        {
            \includegraphics[height=3.05cm]{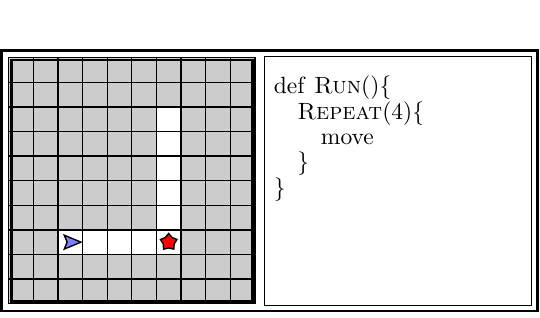}
            \vspace{-4mm}
            \caption{\ours: $(\task^{2}, \code^{2,\ast})$}
            \vspace{-2.9mm}
        }
    \end{subfigure}
    \begin{subfigure}[b]{0.32\textwidth}
        \centering
        {
            \includegraphics[height=3.05cm]{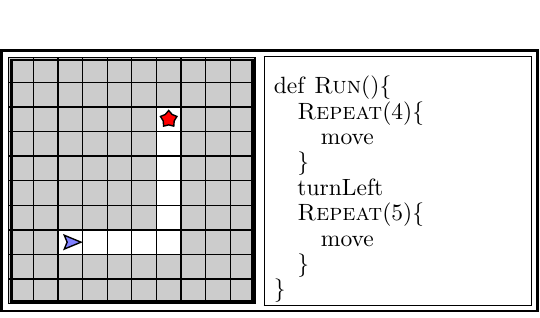}
            \vspace{-4mm}
            \caption{\ours: $(\task^{3}, \code^{3,\ast})$}
            \vspace{-2.9mm}
        }
    \end{subfigure}
    \vspace{1mm}
    \caption{
    \looseness-1Illustration of subtasks synthesized by baseline methods \nodecomp{}, \harddecomp{}, \expnew{}, \expopt{}, and \ours{} for the reference task \hocd{} shown in (a). Each method synthesizes a progression of three subtasks.
    \nodecomp~and \harddecomp~are collectively called \noharddecomp; \expnew~and \expopt~are collectively called \human{}. See \iftoggle{SuppContentOnly}{Section 5 and Appendix~\ref{appendix.sec.userstudy}}{Section~\ref{sec.userstudy} and  Appendix~\ref{appendix.sec.userstudy}} for details.
    }
    \label{fig:appendix.userstudy.hoc}
    \vspace{-5mm}
\end{figure*}

\subsection{Results}\label{sec.userstudy.results}
 We present detailed results in Figure~\ref{table:userstudy} and analyze them w.r.t our RQs.

In total, we had over $500$ participation sessions across two tasks. 
To validate the \textit{usefulness of subtasking (RQ1)}, we compare the success rate on the reference task for methods \nodecomp{} and \ours{}.
We find a $5\%$ increase in the success rate for \ours{}, suggesting the usefulness of subtasks in problem-solving. To investigate the effect of \textit{well-spaced code complexity (RQ2)} in a progression, we compare the success rates for \harddecomp{} and \ours{}. We find that \ours{} outperforms \harddecomp{} by $4.5\%$. This suggests the importance of using progressions with well-spaced code complexity. To investigate the effect of \textit{retaining visual context of the reference task (RQ3)} in the progression, we compare \human{} and \ours{}. We find a $8\%$ increase in success rate for \ours{} compared to \human{}, suggesting the importance of retaining the visual context. Furthermore, we find that \noharddecomp{} also outperforms \human{}. We hypothesize that this is because \human{} synthesizes subtasks that are visually very different from the reference task, possibly making the progression more confusing and leading to lower success rates on the reference task. Also, note the slight difference in the performance of baselines \nodecomp{} and \harddecomp{} (which we together refer to as \noharddecomp{}).  Specifically, we find a slight improvement in performance of \harddecomp{} compared to \nodecomp{}. This is because, the progression of subtasks synthesized by \harddecomp{} have different visual task grids that are minimal modifications of the visual grid of the reference task, while having the same solution code as that of the reference task. However, in the progression of subtasks synthesized by \nodecomp{}, both the code and task grid are exactly the same as that of the reference task. This result also indicates that visual context is an important factor in designing subtasks. 

\paragraph{Limitations and possible extensions.} Next, we discuss a few limitations of our current study. Our study was limited to under $600$ participants given the high costs involved and the reported results are not statistically significant. However, a larger scale user study, with substantially more number of participants, would be needed to further validate the statistical significance of the results. 
Furthermore, we conducted our study with adult novice programmers. In the future, it would be important to conduct longitudinal studies with real students to measure the pedagogical value of our algorithm~\citep{DBLP:journals/jeric/MargulieuxMFR20}. Finally, it would be interesting to evaluate extensions of our approach to more complex block-based programming tasks and domains. 


\section{Conclusions}\label{sec.conclusion}
In this paper, we tackled the problem of synthesizing a progression of subtasks for a given block-based programming task. We proposed a novel algorithm, \ours{}, to automatically synthesize such a progression, using ideas of execution traces and techniques of symbolic execution.
We showcased the utility of \ours{} in improving the problem-solving process of AI agents (neural program synthesizers). Furthermore, we demonstrated the effectiveness of our methodology in assisting novice programmers in solving reference tasks from a popular programming platform.
We have publicly shared the web app used in the study and the implementation of \ours{} to facilitate future work.

%


\textbf{Acknowledgments.} We would like to thank the reviewers for their feedback. Ahana Ghosh acknowledges support by Microsoft Research through its PhD Scholarship Programme.
Funded/Co-funded by the European Union (ERC, TOPS, 101039090). Views and opinions expressed are however those of the author(s) only and do not necessarily reflect those of the European Union or the European Research Council. Neither the European Union nor the granting authority can be held responsible for them.

\clearpage
\bibliographystyle{tmlr}
\bibliography{main}

\begin{thebibliography}{49}
\providecommand{\natexlab}[1]{#1}
\providecommand{\url}[1]{\texttt{#1}}
\expandafter\ifx\csname urlstyle\endcsname\relax
  \providecommand{\doi}[1]{doi: #1}\else
  \providecommand{\doi}{doi: \begingroup \urlstyle{rm}\Url}\fi

\bibitem[Ahmed et~al.(2020)Ahmed, Christakis, Efremov, Fernandez, Ghosh,
  Roychoudhury, and Singla]{DBLP:conf/nips/AhmedCEFGRS20}
Umair~Z. Ahmed, Maria Christakis, Aleksandr Efremov, Nigel Fernandez, Ahana
  Ghosh, Abhik Roychoudhury, and Adish Singla.
\newblock {S}ynthesizing {T}asks for {B}lock-based {P}rogramming.
\newblock In \emph{{N}eur{IPS}}, 2020.

\bibitem[Austin et~al.(2021)Austin, Odena, Nye, Bosma, Michalewski, Dohan,
  Jiang, Cai, Terry, Le, and Sutton]{DBLP:journals/corr/abs-2108-07732}
Jacob Austin, Augustus Odena, Maxwell~I. Nye, Maarten Bosma, Henryk
  Michalewski, David Dohan, Ellen Jiang, Carrie~J. Cai, Michael Terry, Quoc~V.
  Le, and Charles Sutton.
\newblock {P}rogram {S}ynthesis with {L}arge {L}anguage {M}odels.
\newblock \emph{CoRR}, abs/2108.07732, 2021.

\bibitem[Bakker \& Schmidhuber(2004)Bakker and
  Schmidhuber]{bakker2004hierarchical}
Bram Bakker and J{\"u}rgen Schmidhuber.
\newblock {H}ierarchical {R}einforcement {L}earning {B}ased on {S}ubgoal
  {D}iscovery and {S}ubpolicy {S}pecialization.
\newblock In \emph{{I}ntelligent {A}utonomous {S}ystems}, 2004.

\bibitem[Balog et~al.(2017)Balog, Gaunt, Brockschmidt, Nowozin, and
  Tarlow]{DBLP:conf/iclr/BalogGBNT17}
Matej Balog, Alexander~L. Gaunt, Marc Brockschmidt, Sebastian Nowozin, and
  Daniel Tarlow.
\newblock {D}eep{C}oder: {L}earning to {W}rite {P}rograms.
\newblock In \emph{{ICLR}}, 2017.

\bibitem[{B}ishop et~al.(2015){B}ishop, {H}orspool, {X}ie, {T}illmann, and {D}e
  {H}alleux]{bishop2015code}
{J}udith {B}ishop, {R}~{N}igel {H}orspool, {T}ao {X}ie, {N}ikolai {T}illmann,
  and {J}onathan {D}e {H}alleux.
\newblock {C}ode {H}unt: {E}xperience with {C}oding {C}ontests at {S}cale.
\newblock In \emph{{ICSE}}, 2015.

\bibitem[Bunel et~al.(2018)Bunel, Hausknecht, Devlin, Singh, and
  Kohli]{DBLP:conf/iclr/BunelHDSK18}
Rudy Bunel, Matthew~J. Hausknecht, Jacob Devlin, Rishabh Singh, and Pushmeet
  Kohli.
\newblock {L}everaging {G}rammar and {R}einforcement {L}earning for {N}eural
  {P}rogram {S}ynthesis.
\newblock In \emph{ICLR}, 2018.

\bibitem[Burnett \& McIntyre(1995)Burnett and McIntyre]{burnett1995visual}
Margaret~M Burnett and David~W McIntyre.
\newblock {V}isual {P}rogramming.
\newblock \emph{Wiley Encyclopedia of Electrical and Electronics Engineering},
  1995.

\bibitem[Chen et~al.(2019)Chen, Liu, and Song]{DBLP:conf/iclr/ChenLS19}
Xinyun Chen, Chang Liu, and Dawn Song.
\newblock {E}xecution-{G}uided {N}eural {P}rogram {S}ynthesis.
\newblock In \emph{{ICLR}}, 2019.

\bibitem[Codeforces(2022)]{codeforces}
Codeforces.
\newblock {C}odeforces.
\newblock \url{https://codeforces.com/problemset}, 2022.

\bibitem[CodeHS(2022)]{codehscom}
CodeHS.
\newblock {CodeHS.com}: {T}eaching {C}oding and {C}omputer {S}cience.
\newblock \url{https://codehs.com/}, 2022.

\bibitem[Code.org(2022{\natexlab{a}})]{codeorg}
Code.org.
\newblock Code.org: {L}earn {C}omputer {S}cience.
\newblock \url{https://code.org/}, 2022{\natexlab{a}}.

\bibitem[Code.org(2022{\natexlab{b}})]{hourofcode}
Code.org.
\newblock {H}our of {C}ode {I}nitiative.
\newblock \url{https://hourofcode.com/}, 2022{\natexlab{b}}.

\bibitem[Code.org(2022{\natexlab{c}})]{hourofcode_maze}
Code.org.
\newblock {H}our of {C}ode: {C}lassic {M}aze {C}hallenge.
\newblock \url{https://studio.code.org/s/hourofcode}, 2022{\natexlab{c}}.

\bibitem[Decker et~al.(2019)Decker, Margulieux, and
  Morrison]{DBLP:conf/icer/DeckerMM19}
Adrienne Decker, Lauren~E. Margulieux, and Briana~B. Morrison.
\newblock {U}sing the {SOLO} {T}axonomy to {U}nderstand {S}ubgoal {L}abels
  {E}ffect in {CS1}.
\newblock In \emph{{ICER}}, 2019.

\bibitem[Devlin et~al.(2017)Devlin, Uesato, Bhupatiraju, Singh, Mohamed, and
  Kohli]{DBLP:conf/icml/DevlinUBSMK17}
Jacob Devlin, Jonathan Uesato, Surya Bhupatiraju, Rishabh Singh, Abdel{-}rahman
  Mohamed, and Pushmeet Kohli.
\newblock {R}obust{F}ill: {N}eural {P}rogram {L}earning under {N}oisy {I/O}.
\newblock In \emph{{ICML}}, 2017.

\bibitem[Ding et~al.(2014)Ding, LI, and Chuan]{ding2014autonomic}
Xiao Ding, Yi-tong LI, and Shi Chuan.
\newblock {A}utonomic {D}iscovery of {S}ubgoals in {H}ierarchical
  {R}einforcement {L}earning.
\newblock \emph{The Journal of China Universities of Posts and
  Telecommunications}, 2014.

\bibitem[Efremov et~al.(2020)Efremov, Ghosh, and Singla]{edm20-zero-shot}
Aleksandr Efremov, Ahana Ghosh, and Adish Singla.
\newblock Zero-shot {L}earning of {H}int {P}olicy via {R}einforcement
  {L}earning and {P}rogram {S}ynthesis.
\newblock In \emph{EDM}, 2020.

\bibitem[Eysenbach et~al.(2021)Eysenbach, Levine, and
  Salakhutdinov]{DBLP:conf/nips/EysenbachLS21}
Ben Eysenbach, Sergey Levine, and Ruslan Salakhutdinov.
\newblock {R}eplacing {R}ewards with {E}xamples: {E}xample-{B}ased {P}olicy
  {S}earch via {R}ecursive {C}lassification.
\newblock In \emph{{N}eur{IPS}}, 2021.

\bibitem[Ferles et~al.(2017)Ferles, W{\"{u}}stholz, Christakis, and
  Dillig]{DBLP:conf/sigsoft/FerlesWCD17}
Kostas Ferles, Valentin W{\"{u}}stholz, Maria Christakis, and Isil Dillig.
\newblock {F}ailure-{D}irected {P}rogram {T}rimming.
\newblock In \emph{{FSE}}, 2017.

\bibitem[Games(2022)]{blocklygames}
Blockly Games.
\newblock {G}ames for {T}omorrow's {P}rogrammers.
\newblock \url{https://blockly.games/}, 2022.

\bibitem[Ghosh et~al.(2022)Ghosh, Tschiatschek, Devlin, and
  Singla]{aied21_popquiz}
Ahana Ghosh, Sebastian Tschiatschek, Sam Devlin, and Adish Singla.
\newblock {A}daptive {S}caffolding in {B}lock-based {P}rogramming via
  {S}ynthesizing {N}ew {T}asks as {P}op {Q}uizzes.
\newblock In \emph{{AIED}}, 2022.

\bibitem[Gulwani et~al.(2017)Gulwani, Polozov, Singh,
  et~al.]{gulwani2017program}
Sumit Gulwani, Oleksandr Polozov, Rishabh Singh, et~al.
\newblock {P}rogram {S}ynthesis.
\newblock \emph{Foundations and Trends{\textregistered} in Programming
  Languages}, 2017.

\bibitem[HackerRank(2022)]{hackerrank}
HackerRank.
\newblock {H}acker{R}ank.
\newblock \url{https://www.hackerrank.com/}, 2022.

\bibitem[Kiel(2009)]{kiel2009reducing}
Henning Kiel.
\newblock \emph{{R}educing {M}ental {C}ontext {S}witches during {P}rogramming}.
\newblock PhD thesis, RWTH Aachen University, 2009.

\bibitem[King(1976)]{DBLP:journals/cacm/King76}
James~C. King.
\newblock {S}ymbolic {E}xecution and {P}rogram {T}esting.
\newblock \emph{Communications of {ACM}}, 1976.

\bibitem[Korel \& Laski(1988)Korel and Laski]{DBLP:journals/ipl/KorelL88}
Bogdan Korel and Janusz~W. Laski.
\newblock {D}ynamic {P}rogram {S}licing.
\newblock \emph{Information Processing Letters}, 1988.

\bibitem[Laich et~al.(2020)Laich, Bielik, and Vechev]{DBLP:conf/iclr/LaichBV20}
Larissa Laich, Pavol Bielik, and Martin~T. Vechev.
\newblock {G}uiding {P}rogram {S}ynthesis by {L}earning to {G}enerate
  {E}xamples.
\newblock In \emph{{ICLR}}, 2020.

\bibitem[Li et~al.(2022)]{DBLP:journals/corr/abs-2203-07814}
Yujia Li et~al.
\newblock {C}ompetition-{L}evel {C}ode {G}eneration with {A}lphacode.
\newblock \emph{CoRR}, abs/2203.07814, 2022.

\bibitem[Margulieux et~al.(2019)Margulieux, Morrison, and
  Decker]{DBLP:conf/iticse/MargulieuxMD19}
Lauren~E. Margulieux, Briana~B. Morrison, and Adrienne Decker.
\newblock {D}esign and {P}ilot {T}esting of {S}ubgoal {L}abeled {W}orked
  {E}xamples for {F}ive {C}ore {C}oncepts in {CS1}.
\newblock In \emph{Innovation and Technology in Computer Science Education},
  2019.

\bibitem[Margulieux et~al.(2020)Margulieux, Morrison, Franke, and
  Ramilison]{DBLP:journals/jeric/MargulieuxMFR20}
Lauren~E. Margulieux, Briana~B. Morrison, Baker Franke, and Harivololona
  Ramilison.
\newblock {E}ffect of {I}mplementing {S}ubgoals in {C}ode.org's {I}ntro to
  {P}rogramming {U}nit in {C}omputer {S}cience {P}rinciples.
\newblock \emph{{ACM} Transactions on Computing Education}, 2020.

\bibitem[Marwan et~al.(2021)Marwan, Shi, Menezes, Chi, Barnes, and
  Price]{DBLP:conf/edm/Marwan0MCBP21}
Samiha Marwan, Yang Shi, Ian Menezes, Min Chi, Tiffany Barnes, and Thomas~W.
  Price.
\newblock {J}ust a {F}ew {E}xpert {C}onstraints {C}an {H}elp: {H}umanizing
  {D}ata-{D}riven {S}ubgoal {D}etection for {N}ovice {P}rogramming.
\newblock In \emph{{EDM}}, 2021.

\bibitem[McKendree(1990)]{DBLP:journals/hhci/McKendree90}
Jean McKendree.
\newblock {E}ffective {F}eedback {C}ontent for {T}utoring {C}omplex {S}kills.
\newblock \emph{{HCI}}, 1990.

\bibitem[Morrison et~al.(2015)Morrison, Margulieux, and
  Guzdial]{DBLP:conf/icer/MorrisonMG15}
Briana~B. Morrison, Lauren~E. Margulieux, and Mark Guzdial.
\newblock {S}ubgoals, {C}ontext, and {W}orked {E}xamples in {L}earning
  {C}omputing {P}roblem {S}olving.
\newblock In \emph{{ICER}}, 2015.

\bibitem[Morrison et~al.(2016)Morrison, Margulieux, Ericson, and
  Guzdial]{DBLP:conf/sigcse/MorrisonMEG16}
Briana~B. Morrison, Lauren~E. Margulieux, Barbara Ericson, and Mark Guzdial.
\newblock {S}ubgoals {H}elp {S}tudents {S}olve {P}arsons {P}roblems.
\newblock In \emph{{SIGCSE}}, 2016.

\bibitem[Norouzi et~al.(2012)Norouzi, Fleet, and
  Salakhutdinov]{DBLP:conf/nips/0002FS12}
Mohammad Norouzi, David~J. Fleet, and Ruslan Salakhutdinov.
\newblock {H}amming {D}istance {M}etric {L}earning.
\newblock In \emph{{N}eur{IPS}}, 2012.

\bibitem[Pattis et~al.(1995)Pattis, Roberts, and Stehlik]{pattis1995karel}
Richard~E Pattis, Jim Roberts, and Mark Stehlik.
\newblock \emph{{K}arel the {R}obot: {A} {G}entle {I}ntroduction to the {A}rt
  of {P}rogramming}.
\newblock John Wiley \& Sons, Inc., 1995.

\bibitem[Piech et~al.(2015)Piech, Sahami, Huang, and
  Guibas]{DBLP:conf/lats/PiechSHG15}
Chris Piech, Mehran Sahami, Jonathan Huang, and Leonidas~J. Guibas.
\newblock {A}utonomously {G}enerating {H}ints by {I}nferring {P}roblem
  {S}olving {P}olicies.
\newblock In \emph{{L@S}}, 2015.

\bibitem[Polozov et~al.(2015)Polozov, O'Rourke, Smith, Zettlemoyer, Gulwani,
  and Popovic]{DBLP:conf/ijcai/PolozovOSZGP15}
Oleksandr Polozov, Eleanor O'Rourke, Adam~M. Smith, Luke Zettlemoyer, Sumit
  Gulwani, and Zoran Popovic.
\newblock {P}ersonalized {M}athematical {W}ord {P}roblem {G}eneration.
\newblock In \emph{{IJCAI}}, 2015.

\bibitem[Price \& Barnes(2015)Price and Barnes]{Price2015}
Thomas~W. Price and Tiffany Barnes.
\newblock {C}omparing {T}extual and {B}lock {I}nterfaces in a {N}ovice
  {P}rogramming {E}nvironment.
\newblock In \emph{ICER}, 2015.

\bibitem[Price \& Barnes(2017)Price and Barnes]{Price2017PositionPB}
Thomas~W. Price and Tiffany Barnes.
\newblock {P}osition {P}aper: {B}lock-{B}ased {P}rogramming {S}hould {O}ffer
  {I}ntelligent {S}upport for {L}earners.
\newblock \emph{IEEE Blocks and Beyond Workshop}, 2017.

\bibitem[Price et~al.(2017)Price, Zhi, and Barnes]{DBLP:conf/aied/PriceZB17}
Thomas~W. Price, Rui Zhi, and Tiffany Barnes.
\newblock {H}int {G}eneration {U}nder {U}ncertainty: {T}he {E}ffect of {H}int
  {Q}uality on {H}elp-{S}eeking {B}ehavior.
\newblock In \emph{{AIED}}, 2017.

\bibitem[Puri et~al.(2021)Puri, Kung, Janssen, Zhang, Domeniconi, Zolotov,
  Dolby, Chen, Choudhury, Decker, Thost, Buratti, Pujar, Ramji, Finkler,
  Malaika, and Reiss]{DBLP:conf/nips/Puri0JZDZD0CDTB21}
Ruchir Puri, David~S. Kung, Geert Janssen, Wei Zhang, Giacomo Domeniconi,
  Vladimir Zolotov, Julian Dolby, Jie Chen, Mihir~R. Choudhury, Lindsey Decker,
  Veronika Thost, Luca Buratti, Saurabh Pujar, Shyam Ramji, Ulrich Finkler,
  Susan Malaika, and Frederick Reiss.
\newblock {C}ode{N}et: {A} {L}arge-{S}cale {AI} for {C}ode {D}ataset for
  {L}earning a {D}iversity of {C}oding {T}asks.
\newblock In \emph{{N}eur{IPS} {D}atasets and {B}enchmarks}, 2021.

\bibitem[Resnick et~al.(2009)Resnick, Maloney, Monroy-Hern{\'a}ndez, Rusk,
  Eastmond, Brennan, Millner, Rosenbaum, Silver, Silverman,
  et~al.]{DBLP:journals/cacm/ResnickMMREBMRSSK09}
Mitchel Resnick, John Maloney, Andr{\'e}s Monroy-Hern{\'a}ndez, Natalie Rusk,
  Evelyn Eastmond, Karen Brennan, Amon Millner, Eric Rosenbaum, Jay Silver,
  Brian Silverman, et~al.
\newblock {S}cratch: {P}rogramming for {A}ll.
\newblock \emph{{ACM}}, 2009.

\bibitem[Schuster et~al.(2021)Schuster, Kalyan, Polozov, and
  Kalai]{DBLP:conf/nips/SchusterKPK21}
Tal Schuster, Ashwin Kalyan, Alex Polozov, and Adam Kalai.
\newblock {P}rogramming {P}uzzles.
\newblock In \emph{{N}eur{IPS} {D}atasets and {B}enchmarks}, 2021.

\bibitem[Singla \& Theodoropoulos(2022)Singla and
  Theodoropoulos]{edm22-student-synthesis}
Adish Singla and Nikitas Theodoropoulos.
\newblock From \{{S}olution {S}ynthesis\} to \{{S}tudent {A}ttempt
  {S}ynthesis\} for {B}lock-{B}ased {V}isual {P}rogramming {T}asks.
\newblock In \emph{EDM}, 2022.

\bibitem[Singla et~al.(2021)Singla, Rafferty, Radanovic, and
  Heffernan]{DBLP:journals/corr/abs-2107-08828}
Adish Singla, Anna~N. Rafferty, Goran Radanovic, and Neil~T. Heffernan.
\newblock Reinforcement {L}earning for {E}ducation: {O}pportunities and
  {C}hallenges.
\newblock \emph{CoRR}, abs/2107.08828, 2021.

\bibitem[Suh \& Timen(2020)Suh and Timen]{DBLP:journals/corr/abs-2003-10485}
Alexander Suh and Yuval Timen.
\newblock {C}reating {S}ynthetic {D}atasets via {E}volution for {N}eural
  {P}rogram {S}ynthesis.
\newblock \emph{CoRR}, abs/2003.10485, 2020.

\bibitem[Weintrop \& Wilensky(2015)Weintrop and
  Wilensky]{DBLP:conf/acmidc/WeintropW15}
David Weintrop and Uri Wilensky.
\newblock {T}o {B}lock or {N}ot to {B}lock, {T}hat is the {Q}uestion:
  {S}tudents' {P}erceptions of {B}locks-{B}ased {P}rogramming.
\newblock In \emph{{IDC}}, 2015.

\bibitem[Wu et~al.(2019)Wu, Mosse, Goodman, and Piech]{DBLP:conf/aaai/WuMGP19}
Mike Wu, Milan Mosse, Noah~D. Goodman, and Chris Piech.
\newblock {Z}ero {S}hot {L}earning for {C}ode {E}ducation: {R}ubric {S}ampling
  with {D}eep {L}earning {I}nference.
\newblock In \emph{{AAAI}}, 2019.

\end{thebibliography}
\clearpage
\appendix 
{\allowdisplaybreaks
%

\section{List of Appendices}\label{appendix.table-of-contents}
In this section, we provide a brief description of the content in the appendices of the paper.   
\begin{itemize}
\item Appendix~\ref{appendix.sec.problemsetup} provides a table of notations used throughout the paper.
\item Appendix~\ref{appendix.sec.model} provides additional details about our synthesis algorithm \synsubtasks{}. (\iftoggle{SuppContentOnly}{Section 3}{Section~\ref{sec.algorithm}})
\item Appendix~\ref{appendix.sec.userstudy} provides additional details about our study with novice human programmers. (\iftoggle{SuppContentOnly}{Section 5}{Section~\ref{sec.userstudy}})
\end{itemize}
\section{Table of Notations (\iftoggle{SuppContentOnly}{Section 2}
{Section~\ref{sec.problemsetup}})}\label{appendix.sec.problemsetup}
See Figure~\ref{appendix.notation.table} for the complete list of notations.
\begin{figure*}[h!]
\scalebox{0.85}{
\renewcommand{\arraystretch}{1.30}
        \begin{tabular}{r|l}
    		\toprule
    		\textbf{Notation} & \textbf{Description}
    		\\
    		\toprule
    		\task & Task identifier
                \\
                $\task_\textnormal{n}$ & Number of visual grids in task \task
                \\
                $\task_\textnormal{vis}$ & The set of visual grids of task \task
                \\
                $\task_\textnormal{store}$ & The types of code blocks available to solve the task \task
                \\
                $\task_\textnormal{size}$ & The maximum number of code blocks allowed in the solution code of task \task
                \\
                $\mathbb{T}$ & Task space
                \\
                $\mathcal{F}^\mathbb{T}_\textnormal{complex}: \mathbb{T} \rightarrow \mathbb{R}$ & Function to measure the complexity of a task
                \\
                $\mathcal{F}^\mathbb{T}_\textnormal{diss}: \mathbb{T} \times \mathbb{T} \rightarrow \mathbb{R}$ & Function to measure the dissimilarity between two tasks
                \\
                \code & Code indentifier
                \\
                $\code_\textnormal{depth}$ & Depth of the Abstract Syntax Tree (AST) of a code \code
                \\
                $\code_\textnormal{size}$ & Number of code blocks in code \code
                \\
                 $\code_\textnormal{blocks}$ & Types of code blocks in code \code
                 \\
                 $\mathbb{C}$ & Code space
                 \\
                  $\mathcal{F}^\mathbb{C}_\textnormal{complex}: \mathbb{C} \rightarrow \mathbb{R}$ & Function to measure the complexity of a code
                  \\

                  $\mathbb{C}_\textnormal{T}$ & Set of all solution codes of task \task
                  \\
                  $K$ & Fixed budget on the number of subtasks for a reference task
                  \\
                  $\omega$ & A specific progression of subtasks for a reference task
                  \\
                  $\Omega$ & Set of all progressions of subtasks for a reference task
                  \\
                  $\mathcal{F}^{\Omega}_\textnormal{complex}$ & Function to measure the complexity of a progression of subtasks for a reference task
                  \\
                  $\executiontracetask^{\task}$ & State of the first visual grid of task \task~at time-step $\tau$ 
                  \\
                  $\executiontracetask^{\code}$ & Partial code obtained from code \code, after it is executed till time-step $\tau$
                  \\
                  $\executiontrace$ & Execution trace of a code on a single grid of a task; obtained after Stage $1$ of \synsubtasksingle
                  \\
                  $\executiontracefiltered$ & Filtered execution trace of a code on a single grid of a task; obtained after Stage $2$ of \synsubtasksingle
                  \\
                  $\executiontracesymexec$ & Modified execution trace of a code on a single grid of a task; obtained after Stage $3$ of \synsubtasksingle
                  \\
                  $\Sigma^\textnormal{p}$ & set of all permutations of $\{1,2,\ldots,p\}$
                  \\
                  $\sigma$ & A specific permutation from the set of permutations $\Sigma^\textnormal{p}$
                  \\
    		\bottomrule
\end{tabular}
}
\caption{\label{appendix.notation.table} Table of notations}
\end{figure*}

\section{Our Synthesis Algorithm (\iftoggle{SuppContentOnly}{Section 3}{Section~\ref{sec.algorithm}})}\label{appendix.sec.model}
In this section, we discuss additional details of our algorithm \synsubtasks. We begin by discussing additional details of the procedure \synsubtasksmulti{}. After that, we discuss additional details of our algorithm to synthesize a progression of subtasks for a reference task with a single visual grid (\synsubtasksingle). Finally, we present the detailed algorithm \synsubtasks{} and share details of our implementation with pointers to specific code files.

\subsection{\synsubtasksmulti{}: Additional Details (\iftoggle{SuppContentOnly}{Section 3.1}{Section~\ref{sec.algorithm.multigrid}})}\label{appendix.model.grids}
Next, we present additional details of \synsubtasksmulti{}, discussed in Section \iftoggle{SuppContentOnly}{3.1}{~\ref{sec.algorithm.multigrid}}. Specifically, we discuss the optimization strategies one can adopt for solving \iftoggle{SuppContentOnly}{Equation 2}{Equation~\ref{eq:problemstatement.1}} when the number of visual grids $\task^\textnormal{ref}_\textnormal{n}$, of a task $\task^\textnormal{ref}$ is large. In this case, we can decide the sequence of subtasks generated in the final progression using a greedy strategy. We can use the degree of code coverage to generate the sequence. We can begin with grids corresponding to maximum code coverage of the solution code, and sequentially add the remaining grids.

When $\task^\textnormal{ref}_\textnormal{n}$ is small (e.g., $\leq 6$ in domains we are considering -- such as the task shown in \iftoggle{SuppContentOnly}{Figure 2}{Figure~\ref{fig:intro.multigrid.og}} in \iftoggle{SuppContentOnly}{Section 1}{Section~\ref{sec.intro}}), one can optimize for \iftoggle{SuppContentOnly}{Equation 2}{Equation~\ref{eq:problemstatement.1}} by enumerating all possible elements of \progressionset.

\subsection{\synsubtasksingle{}: Additional Details (\iftoggle{SuppContentOnly}{Section 3.2}{Section~\ref{sec.algorithm.singlegrid}})}\label{appendix.model.singlegrid}
Next, we present additional details for each of the four stages of \synsubtasksingle, for a reference task $\task^\textnormal{ref}$ and its solution code $\code^{\textnormal{ref},\ast}$, where $\task^\textnormal{ref}_\textnormal{n} = 1$.

\paragraph{Stage 1: Execution trace of code on the single visual grid (\iftoggle{SuppContentOnly}{Figure 3a}{Figure  ~\ref{fig:model2.hoc.stage1}}).}
In this stage, we obtain the full execution trace $\executiontrace(\task^\textnormal{ref}, \code^{\textnormal{ref},\ast})$ of the solution code $\code^{\textnormal{ref},\ast}$ on the visual grid $\task^\textnormal{ref}_\textnormal{vis,1}$. We obtain the trace by instrumenting a Karel interpreter (which is used for executing Karel programs) as it executes $\code^{\textnormal{ref},\ast}$. 

\looseness-1\paragraph{Stage 2: Post-processing the trace based on code validity and code quality (\iftoggle{SuppContentOnly}{Figure 3b}{Figure 3b~\ref{fig:model2.hoc.stage2}}).}
In this stage we post-process the execution trace obtained in Stage 1, based on code validity and code quality, to obtain $\executiontracefiltered(\task^\textnormal{ref}, \code^{\textnormal{ref},\ast})$. Specifically, we filter invalid codes from the trace whose code command sequence does the following: The code command sequence terminates on a command/code block that occurs in the middle of a loop/conditional body of $\code^{\textnormal{ref},\ast}$. See \iftoggle{SuppContentOnly}{Figure 3b}{Figure~\ref{fig:model2.hoc.stage2}}, where the code at step $\tau=8$ is filtered as its corresponding sequence of code commands in \iftoggle{SuppContentOnly}{Figure 3a}{Figure~\ref{fig:model2.hoc.stage1}} terminates on code block \DSLMove{}, which is in the middle of the body of the \DSLRepeatUntil{} construct of the solution code $\code^{\textnormal{ref},\ast}$ (shown in \iftoggle{SuppContentOnly}{Figure 1a}{Figure~\ref{fig:intro.hoc.og}}). A loop refers to the code constructs, \DSLRepeatUntil{}, \DSLWhile{} and \DSLRepeat{}. A conditional refers to the code constructs, \DSLIf{} and \DSLIfElse{}.

From the remaining code commands we generate concrete codes. To generate the concrete codes, we use compiler information and the AST structure of the code. We instrument the compiler to maintain information on which branches of the AST are executed in the current sequence of code commands. Using this information, we can directly convert the sequence of commands into code. However, to maintain code quality, we insert a loop construct in the code only when the body of the loop is executed more than once in the code command sequence; when there are conditional branches inside the loop body, we add the loop construct when at least one of the branches is executed more than once. For the other codes, we retain the unrolled loop body, without the loop construct.

\paragraph{Stage 3: Modifying grids in the trace via symbolic execution (\iftoggle{SuppContentOnly}{Figure 3c}{Figure~\ref{fig:model2.hoc.stage3}}).}
In this stage, we synthesize task grids for each of the codes from the sequence obtained in Stage 2. We denote the final sequence after this stage as $\executiontracesymexec(\task^\textnormal{ref}, \code^{\textnormal{ref},\ast})$. We obtain high quality visual task grids for each of the codes using symbolic execution techniques and constraint solvers. Specifically, during symbolic execution we make minimal modifications to the visual grids of the subtasks w.r.t. the reference task $\task^\textnormal{ref}$, to generate valid tasks. Next, we describe the key ideas behind this stage.

In the sequence \executiontracefiltered{} obtained from Stage 2, some of the visual grid and code pairs are inconsistent (i.e., the code is not a valid solution for the visual grid) as the code execution is terminated prematurely at those steps. For example, in \iftoggle{SuppContentOnly}{Figure 3}{Figure~\ref{fig:model2.hoc}}, step $\tau=12$ shows a scenario where the code execution has terminated while the avatar (purple dart) has not reached the goal (red star) yet on the visual grid. To resolve this inconsistency, we use the trace on the visual grid as a specific path in the symbolic execution of the corresponding code. The trace determines the boolean values of the conditionals (like \DSLIf{}, \DSLIfElse{}, \DSLWhile{}) of the code during symbolic execution. Combined with the trajectory of the avatar's locations, this trace gives us a set of constraints over a subset of grid cells of the visual task grid. If an assignment  satisfying all the constraints does not exist, we invalidate the code and it is eliminated from the sequence along with the visual grid. Otherwise, we use the assignment values for the constrained grid cells. For the grid cells without any constraints, we retain their values from the visual grid of the reference task to ensure minimal modification (see \iftoggle{SuppContentOnly}{Figure 3c}{Figure~\ref{fig:model2.hoc.stage3}}). Limiting the computation to the execution trace and the avatar's trajectory on the visual grid not only allows us to retain the visual context of the reference task, but also makes this operation computationally less expensive.

\paragraph{Stage 4: Generating subtasks via subsequence selection.}
In this final stage, we obtain a progression of $K'$ subtasks, from the set of all subsequences of length $K'$ from \executiontracesymexec{}. We denote this set of sequences as $\progressionset(\task^\textnormal{ref},\code^{\textnormal{ref},\ast}, K')$. Using 
\progressionset, we optimize for \iftoggle{SuppContentOnly}{Equation 2}{Equation~\ref{eq:problemstatement.1}} to obtain our final progression. Specifically, we apply techniques of dynamic programming to optimize for \iftoggle{SuppContentOnly}{Equation 2}{Equation~\ref{eq:problemstatement.1}} and select a sequence of $K'$ elements from \executiontracesymexec{}. This way, we do not need to enumerate over all elements of \progressionset. While applying dynamic programming, we ensure that the following desirable properties are part of our synthesized progression of tasks: (i) $\text{maximize}~\Sigma_{k \in \{1,\ldots,K'\}} \FTaskquality(\task^k)$; (ii) $\text{minimize}~\Sigma_{k\in\{1,\ldots,K'\}} \FDissimilarity(\task^k, \task^\textnormal{ref})$; (iii) subtasks in the progression are diverse from each other.

\subsection{\synsubtasks{}: Detailed Algorithm (\iftoggle{SuppContentOnly}{Section 3.3}{Section~\ref{sec.algorithm}})}\label{appendix.model.alg}
The detailed algorithm to synthesize a progression of subtasks for a given reference task $\task^\textnormal{ref}$ and its solution code $\code^{\textnormal{ref},\ast}$ is presented in Algorithm~\ref{alg:synthesis}. Note that, we present only the pseudo-code of our algorithm. Our implementation optimizes the algorithm further to avoid redundant computations and run-time overheads.

\paragraph{A few important points about our algorithm:} 
\begin{itemize}
\item In our implementation of the code complexity function $\FCodecomplexity = \kappa * \code_\textnormal{depth} + \code_\textnormal{size}$ (see \iftoggle{SuppContentOnly}{Section 2.1}{Section~\ref{sec:problemstatement.prelims}}), we set $\kappa = 1000$.
\item The time-complexity of our dynamic programming routine is $O(K(M^{SE})^{3})$, indicating that our algorithm is linear w.r.t. the number of subtasks $K$. Here, $M^{SE}$ is bounded by the length of the execution trace of the solution code $\code^{\ast}$ on task $\task$ (See Stage 1 of \synsubtasksingle{} in \iftoggle{SuppContentOnly}{Section 3.2}{Section~\ref{sec.algorithm.singlegrid}}).
\item In our procedure, the number of subtasks $K$ provides a trade-off between the number of subtasks that a student must solve, and the complexity jumps between consecutive subtasks. In practice, one can assume that a teacher provides an upper and lower bound on $K$, as well as a bound on the maximum complexity jump in the sequence of subtasks. Our algorithm could be extended to find the minimal $K$  satisfying these constraints. For the reference tasks shown in \iftoggle{SuppContentOnly}{Figure 1a}{Figure~\ref{fig:intro.hoc.og}} and \iftoggle{SuppContentOnly}{Figure 2a}{Figure~\ref{fig:intro.multigrid.og}} (in \iftoggle{SuppContentOnly}{Section 1}{Section~\ref{sec.intro}}), we set $K'=3$ for our single grid decomposition procedure \synsubtasksingle{}.  For the multi-grid Karel task, Stairway (shown in \iftoggle{SuppContentOnly}{Figure 2a}{Figure~\ref{fig:intro.multigrid.og}}), which had $\task^\textnormal{ref}_\textnormal{n} = 3$ visual grids, this resulted in $K' + \task^\textnormal{ref}_\textnormal{n} -1 = 3 + 3 -1 = 5$ subtasks.
\end{itemize}


\begin{algorithm*}[!h]
    \caption{\synsubtasks}
    \begin{algorithmic}[1]
        \Function{\synsubtasks}{$\task^\textnormal{ref}, \code^\textnormal{ref,*}, K$}\label{alg:appendix.synthesis.combined}
        \State \textbf{Initialize: } $\progressionset \leftarrow \{\}$; $\Sigma^{\task^\textnormal{ref}_\textnormal{n}} \leftarrow \text{set of all permutations of }\{1,2,\ldots,\task^\textnormal{ref}_\textnormal{n}\}$
        \For{$\sigma \in \Sigma^{\task^\textnormal{ref}_\textnormal{n}}$}   
        \State $\code^\textnormal{single,*} \leftarrow \reducedcode(\{\task^\textnormal{ref}_{\textnormal{vis},\sigma_1}\}; \task^\textnormal{ref}, \code^\textnormal{ref,*})$ 
        \State $\task^\textnormal{single} := (1, \{\task^\textnormal{ref}_{\textnormal{vis,}\sigma_{1}}\}, \code^\textnormal{single,*}_\textnormal{blocks}, \code^\textnormal{single,*}_\textnormal{size})$
           \State $\progression1 \leftarrow$  $\synsubtasksingle(\task^\textnormal{single}, \code^{\textnormal{single},\ast}, K')$ where $K' = K-\task^\textnormal{ref}_\textnormal{n} + 1$
           \State $\progression2 \leftarrow \synsubtasksmulti(\task^\textnormal{ref}, \code^{\textnormal{ref},\ast}, \sigma)$, i.e., progression for a given permutation $\sigma$ %
             \State $\progression{} \leftarrow$ Concatenate $\progression1$ with $\progression2$ (after removing the common element)
            \State Add \progression~to \progressionset
        \EndFor{}
        \State $\progression^{*} = \arg \min_{\progression \in \progressionset} \FProgressioncomplexity(\progression; \task^\textnormal{ref}, \code^\textnormal{ref,*},K)$  as per \iftoggle{SuppContentOnly}{Equation 2}{Equation~\ref{eq:problemstatement.1}}
        \State \textbf{Return $\progression^{*} \in \progressionset$}  
        \EndFunction
      
        \text{ }
        \text{ }
        \Function{\synsubtasksingle}{$\task^\textnormal{single}, \code^\textnormal{single,*}, K'$}\label{alg:synthesis.single}
            \State [Stage 1] Obtain the execution trace $\executiontrace(\task^\textnormal{single}, \code^\textnormal{single,*})$
            \State [Stage 2] Post-process trace based on code validity and quality to obtain  $\executiontracefiltered(\task^\textnormal{single}, \code^\textnormal{single,*})$
            \State [Stage 3] Modify grids in trace via symbolic execution to obtain  $\executiontracesymexec(\task^\textnormal{single}, \code^\textnormal{single,*})$
            \State [Stage 4] Define $\progressionset$ as the set of all $K'$-length subsequences of \executiontracesymexec
            \State [Stage 4] $\progression^{*} = \arg \min_{\progression \in \progressionset} \FProgressioncomplexity(\progression; \task^\textnormal{single}, \code^\textnormal{single,*},K')$ as per \iftoggle{SuppContentOnly}{Equation 2}{Equation~\ref{eq:problemstatement.1}}
            \State \textbf{Return $\progression^{*} \in \progressionset$}  
        \EndFunction
        
        \text{ }
        \text{ }
        \Function{\synsubtasksmulti}{$\task^\textnormal{ref}, \code^\textnormal{ref,*}, \sigma$} \label{alg:synthesis.multi}
        \State \textcolor{orange}{// When a fixed $\sigma$ is provided as input (the case when the procedure is invoked from \ours{})}
        \State For $\sigma$, we define a sequence of $\task^{\textnormal{ref}}_{\textnormal{n}}$ tasks as $\progression^\sigma$ 
        \State We define the $k$-th task in $\progression^\sigma$ as follows:
        \Indent
        \State $\code \leftarrow \reducedcode(\{\task^{\textnormal{ref}}_{\textnormal{vis},\sigma_{i}}\}_{i=1,\ldots,k}; \task^\textnormal{ref}, \code^\textnormal{ref,*})$
        \State $\task := (k, \{\task^{\textnormal{ref}}_{\textnormal{vis},\sigma_{i}}\}_{i=1,\ldots,k}, \code_\textnormal{blocks}, \code_\textnormal{size})$
        \State $k$-th subtask and solution code $ := (\task, \code)$
        \EndIndent
        \State \textbf{Return} $\progression^\sigma$
        \State \textcolor{orange}{// When there is no $\sigma$ as input (the case when this procedure is used separately as discussed in \iftoggle{SuppContentOnly}{Section 3.1}{Section~\ref{sec.algorithm.multigrid}})}
        \State \textbf{Initialize: } $\Sigma^{\task^\textnormal{ref}_\textnormal{n}} \leftarrow \text{set of all permutations of }\{1,2,\ldots,\task^\textnormal{ref}_\textnormal{n}\}$; $\progressionset \leftarrow \{\}$
         \For{$\sigma \in \Sigma^{\task^\textnormal{ref}_\textnormal{n}}$}   
            \State For $\sigma$, we define a sequence of $\task^{\textnormal{ref}}_{\textnormal{n}}$ tasks as $\progression^\sigma$ 
            \State We define the $k$-th task in $\progression^\sigma$ as follows:
            \Indent
            \State $\code \leftarrow \reducedcode(\{\task^{\textnormal{ref}}_{\textnormal{vis},\sigma_{i}}\}_{i=1,\ldots,k}; \task^\textnormal{ref}, \code^\textnormal{ref,*})$
            \State $\task := (k, \{\task^{\textnormal{ref}}_{\textnormal{vis},\sigma_{i}}\}_{i=1,\ldots,k}, \code_\textnormal{blocks}, \code_\textnormal{size})$
            \State $k$-th subtask and solution code $ := (\task, \code)$
            \EndIndent
        \State Add $\progression^\sigma$ to $\progressionset$
        \EndFor{}
         \State $\progression^{*} = \arg \min_{\progression \in \progressionset} \FProgressioncomplexity(\progression; \task^\textnormal{ref}, \code^\textnormal{ref,*},\task^\textnormal{ref}_\textnormal{n})$ as per \iftoggle{SuppContentOnly}{Equation 2}{Equation~\ref{eq:problemstatement.1}}
        \State \textbf{Return} $\progression^{*} \in \progressionset$  
        \EndFunction
    \end{algorithmic}
    \label{alg:synthesis}
\end{algorithm*}

\subsection{\synsubtasks{}: Implementation files}\label{appendix.sec.model.implementation}
We present three demo scripts in our supplementary code files, which can be executed to obtain progression of subtasks synthesized by \synsubtasks{} for reference tasks \hocd{} (see Figure~\ref{fig:appendix.hoc.og}), \hocg{} (see \iftoggle{SuppContentOnly}{Figure 1a}{Figure~\ref{fig:intro.hoc.og}}) and \textsc{Stairway} task from \emph{CodeHS.com} based on the \emph{Karel programming environment} (see \iftoggle{SuppContentOnly}{Figure 2a}{Figure~\ref{fig:intro.multigrid.og}}). The scripts are located in the folder ``code/algorithm\_progresssyn\_demo'' and are as follows:
\begin{itemize}
    \item Demo for \hocd{}: \textcode{progressyn\_hoc08.py}
    \item Demo for \hocg{}: \textcode{progressyn\_hoc16.py}
    \item Demo for Karel \textsc{Stairway}: \textcode{progressyn\_stairway.py}
\end{itemize}

Specifically, in our supplementary code folder ``code/algorithm\_progressyn/subtasking/'',  the following functions correspond to our synthesis algorithms:
\begin{itemize}
    \item \synsubtasks{} is implemented in the function \textcode{progressyn()}
    \item \synsubtasksingle{} is implemented in the function \textcode{progressyn\_single()}
    \item \synsubtasksmulti{} is implemented in the function \textcode{progressyn\_grids()}
\end{itemize}

\clearpage


\begin{figure*}[t!]
\centering
        \scalebox{0.7}{
	    \renewcommand{\arraystretch}{1.50}
        \begin{tabular}{r||rrrr||rrrr||rrrr}
    			\toprule
    			Method & \multicolumn{4}{c||}{Average Quality} & \multicolumn{4}{c||}{Normalized Task Diversity} &
    			\multicolumn{4}{c}{Maximum Complexity Jump}
    			\\
    			 & All  & \hocd & \hocg & 
                 $\textnormal{K}_\textnormal{sgStair}$ &
    			 All  & \hocd & \hocg &
                 $\textnormal{K}_\textnormal{sgStair}$ &
    			 All  & \hocd & \hocg & 
                 $\textnormal{K}_\textnormal{sgStair}$
    			 \\
    			\toprule
    			\default & $1.$ & $1.$ & $1.$ & 
                    $1.$
    			& $0.$ & $0.$ & $0.$ &
       $0.$
    			& $1405.4$ & $1005$ & $2004$ & 
       $2007$
    			\\
     			\hline
    			 \noharddecomp & $1.$ & $1.$ & $1.$ &
        $1.$
    			& $0.75$ & $0.75$ & $0.75$ &
       $0.75$
    			& $1405.4$ & $1005$ & $2004$ &
       $2007$
                \\
                \human & $1.$ & $1.$ & $1.$ &
                $1.$
                & $1.21$ & $1.1$ & $1.13$ &
                $1.75$
                & $1000.$ & $1002$ & $1002$ &
                $1001$
                \\ \hline
    			\ours & $1.$ & $1.$ & $1.$ &
       $1.$
    		    & $1.7$ & $1.5$ & $1.5$ &
          $2.$
    		    & $999.2$ & $1000$ & $1002$ &
          $1001$
                \\
    			\bottomrule
        \end{tabular}
        }
    \vspace{1.5mm}
\caption{Validation of metrics task quality, task dissimilarity, task diversity and task complexity for progression of subtasks synthesized by different algorithms. Higher values of average quality and normalized task diversity are better while lower values of maximum complexity jump are better; see details in Appendix~\ref{appendix.userstudy_validation}.}
\label{fig:appendix_mc_validation}
\vspace{-6mm}
\end{figure*}

\section{Assisting Novice Human Programmers (\iftoggle{SuppContentOnly}{Section 5}{Section~\ref{sec.userstudy}})}\label{appendix.sec.userstudy}
\looseness-1In this section, we present additional results evaluating different subtasking algorithms w.r.t the desirable properties of a progression (as described in \iftoggle{SuppContentOnly}{Section 2.2}{Section~\ref{sec.problemsetup}}). We also provide additional details about the web app used for the user study.

\subsection{Validation: Additional Results}\label{appendix.userstudy_validation}
In this section, we discuss the importance of properties: task quality, task dissimilarity, task diversity, and well-spaced code complexity (as described in \iftoggle{SuppContentOnly}{Section 2.2}{Section~\ref{sec.problemsetup}}), for synthesizing a good progression of subtasks for a given reference task. In particular, properties of task quality, task dissimilarity and task diversity were added to our optimization problem (\iftoggle{SuppContentOnly}{Equation 2}{Equation~\ref{eq:problemstatement.1}}) to ensure that we have a single optimal sequence of subtasks for a given reference task. These properties were specifically used for tie-breaking. We designed our baselines (\nodecomp{}, \harddecomp{}, \human{}) in a manner that violated one or more desirable properties of the subtasks. Specifically,
\begin{itemize}
    \item \nodecomp{} violates properties: (i) task diversity and (ii) well-spaced task complexity.
    \item \harddecomp{} violates properties: (i) well-spaced task complexity because the solution codes of all the subtasks were the same.
    \item \human{} violates the following properties: (i) task dissimilarity w.r.t reference task.
\end{itemize}

So, the performance of these three baselines and \ours{} with novice human programmers (discussed in \iftoggle{SuppContentOnly}{Section 5}{Section~\ref{sec.userstudy}}) highlights the degree to which these properties are important in achieving the overall goal of improving performance on the reference tasks. 

We also provide the exact values of these properties for the progression of subtasks synthesized by each of the methods (\default{}, \noharddecomp{}, \human{} and \ours{}) for reference tasks \hocd{}, \hocg{}, and the single-grid variant of \emph{Karel programming environment} based task \textsc{Stairway}\footnote{The multi-grid variant of \emph{Karel programming environment} based task \textsc{Stairway} is illustrated in Figure~\ref{fig:intro.multigrid.og}.} (referred to as $\textnormal{K}_\textnormal{sgStair}$) in Figure~\ref{fig:appendix_mc_validation}. Specifically, in Figure~\ref{fig:appendix_mc_validation} we present the following properties:
\begin{itemize}
    \item Average Quality:  This measures the average quality of subtasks in the final progression. In our implementation, we used the definition of task quality from~\citet{DBLP:conf/nips/AhmedCEFGRS20}; in particular, we used a binary indicator of quality to be 1 if it is above a threshold.
    \item Normalized Task Diversity: We define the normalized task diversity in the final progression of subtasks as, $\frac{2}{K-1}. \frac{\text{Task dissimilarity between subtasks}}{\text{Task dissimilarity between subtasks and the reference task}}$ where,
    \begin{itemize}
        \item $K$ = Number of subtasks in the final progression
        \item Task dissimilarity between subtasks = $\Sigma_{i=1,\ldots,K}\Sigma_{j=1,\ldots,i} \FDissimilarity(\task^i, \task^j)$
        \item Task dissimilarity between subtasks and the reference task = $\Sigma_{k=1,\ldots,K}\FDissimilarity(\task^{k}, \task^\textnormal{ref})$
    \end{itemize}
    \item Maximum Complexity Jump: This measures the maximum difference in task complexity in the final progression of subtasks, where, task complexity is given by $\FComplexity^{\taskspace}(\task) = 1000*\code^{\task, \ast}_\textnormal{depth} + \code^{\task, \ast}_\textnormal{size}$.
\end{itemize}
From Figure~\ref{fig:appendix_mc_validation} we find that compared to all the baselines, \ours{} achieves higher task diversity and has well-spaced task complexity (minimal value of maximum complexity jump in final progression of subtasks). Note that, baseline \default{} has only one task-code pair in the progression which is the same as the reference task and its solution code. Hence, we set its normalized diversity score to $0$.


\begin{figure*}[t!]
\centering
    \begin{subfigure}[t]{0.48\textwidth}
    \centering
        \frame{\includegraphics[trim={0.2cm 0.2cm 0.5cm 0.1cm},clip,height=4.6cm]{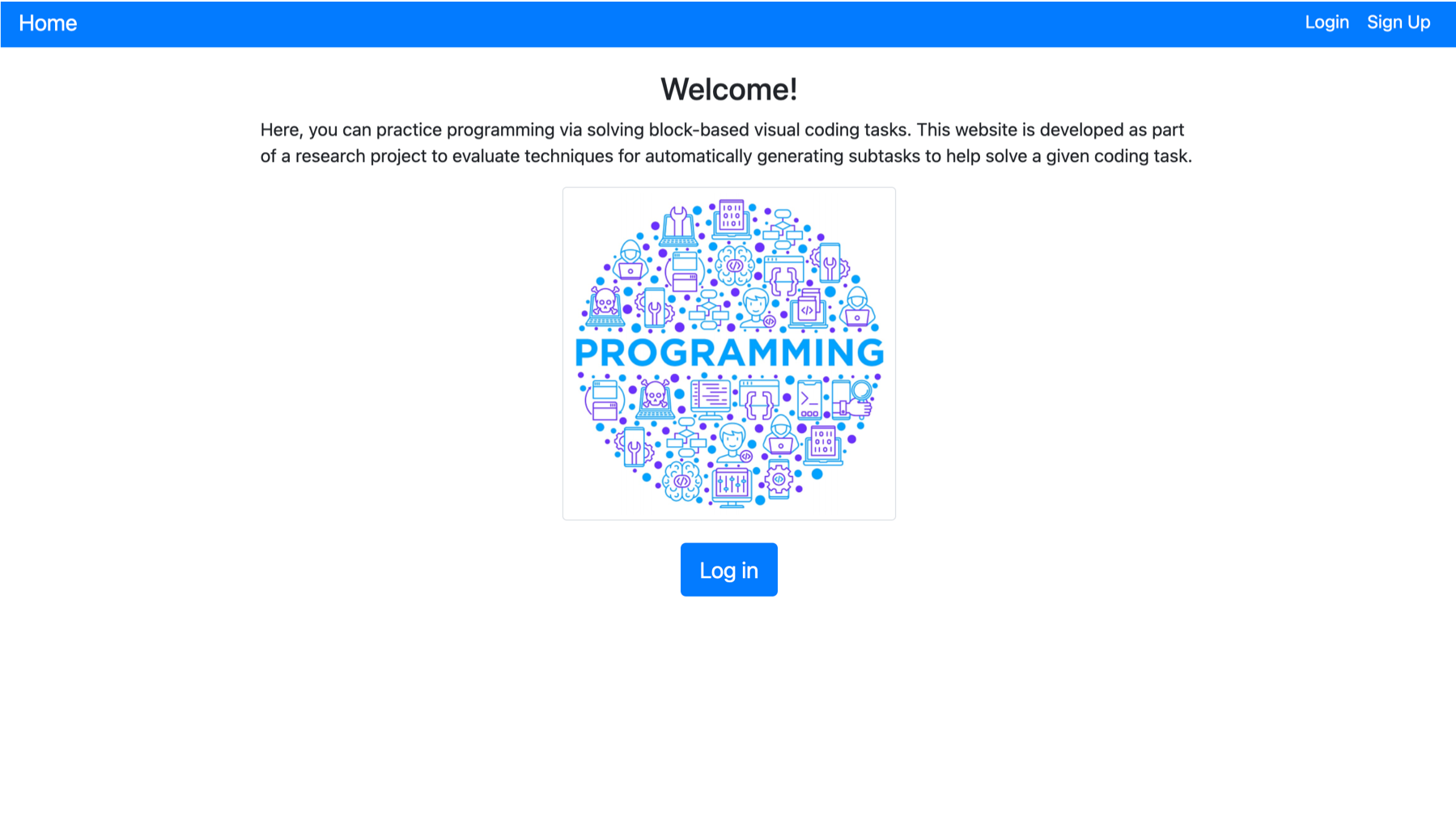}}
    \caption{Login and welcome page}
    \end{subfigure}
    \hspace{3mm}
     \begin{subfigure}[t]{0.48\textwidth}
     \centering
        \frame{\includegraphics[trim={0.2cm 0.2cm 0.5cm 0.1cm},clip,height=4.6cm]{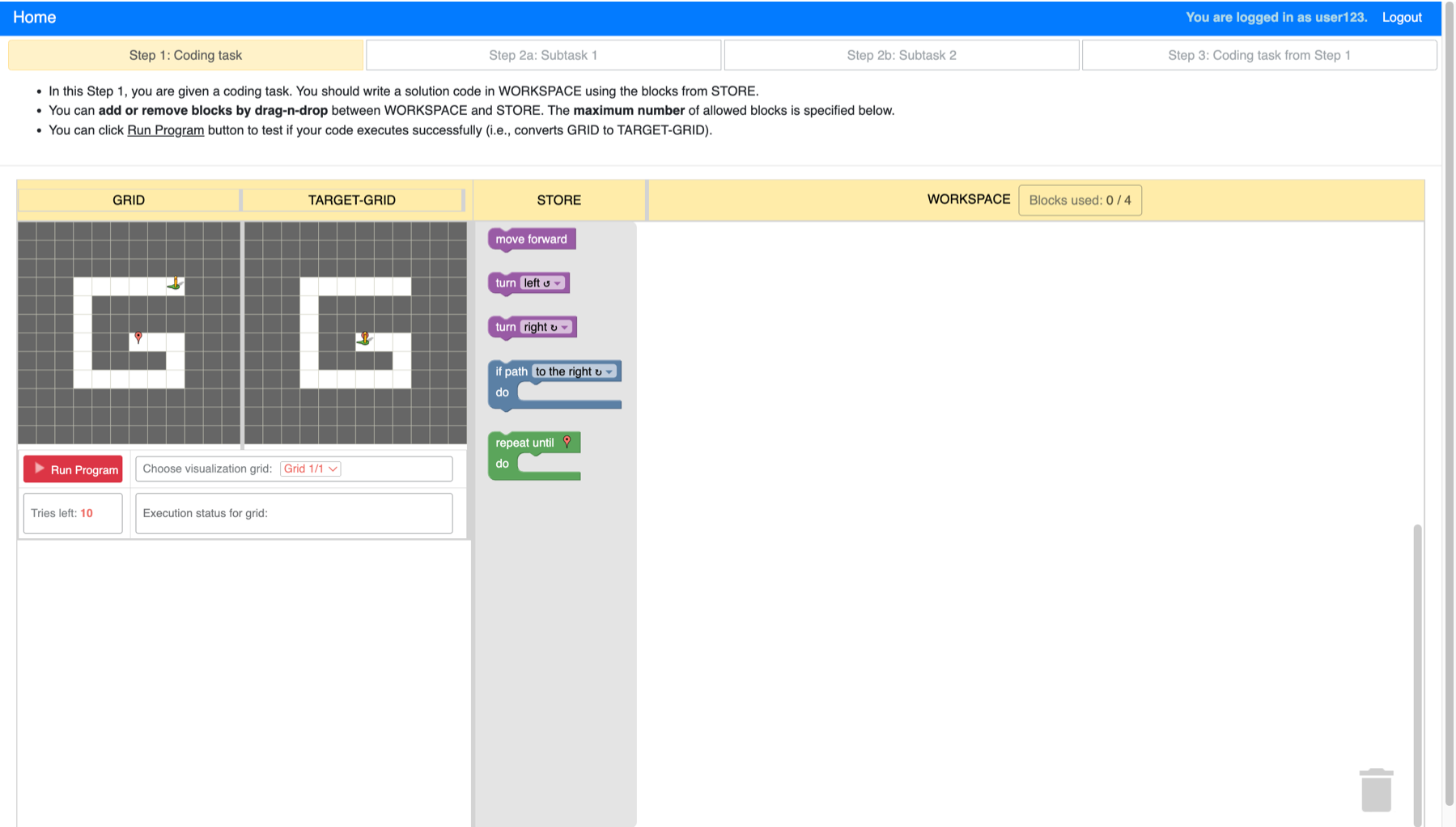}}
    \caption{Step $1$: Introducing reference task $\task^\textnormal{ref}$}
    \end{subfigure}
    \\ \vspace{4mm}
     \begin{subfigure}[t]{0.48\textwidth}
     \centering
        \frame{\includegraphics[trim={0.2cm 0.2cm 0.5cm 0.1cm},clip,height=4.6cm]{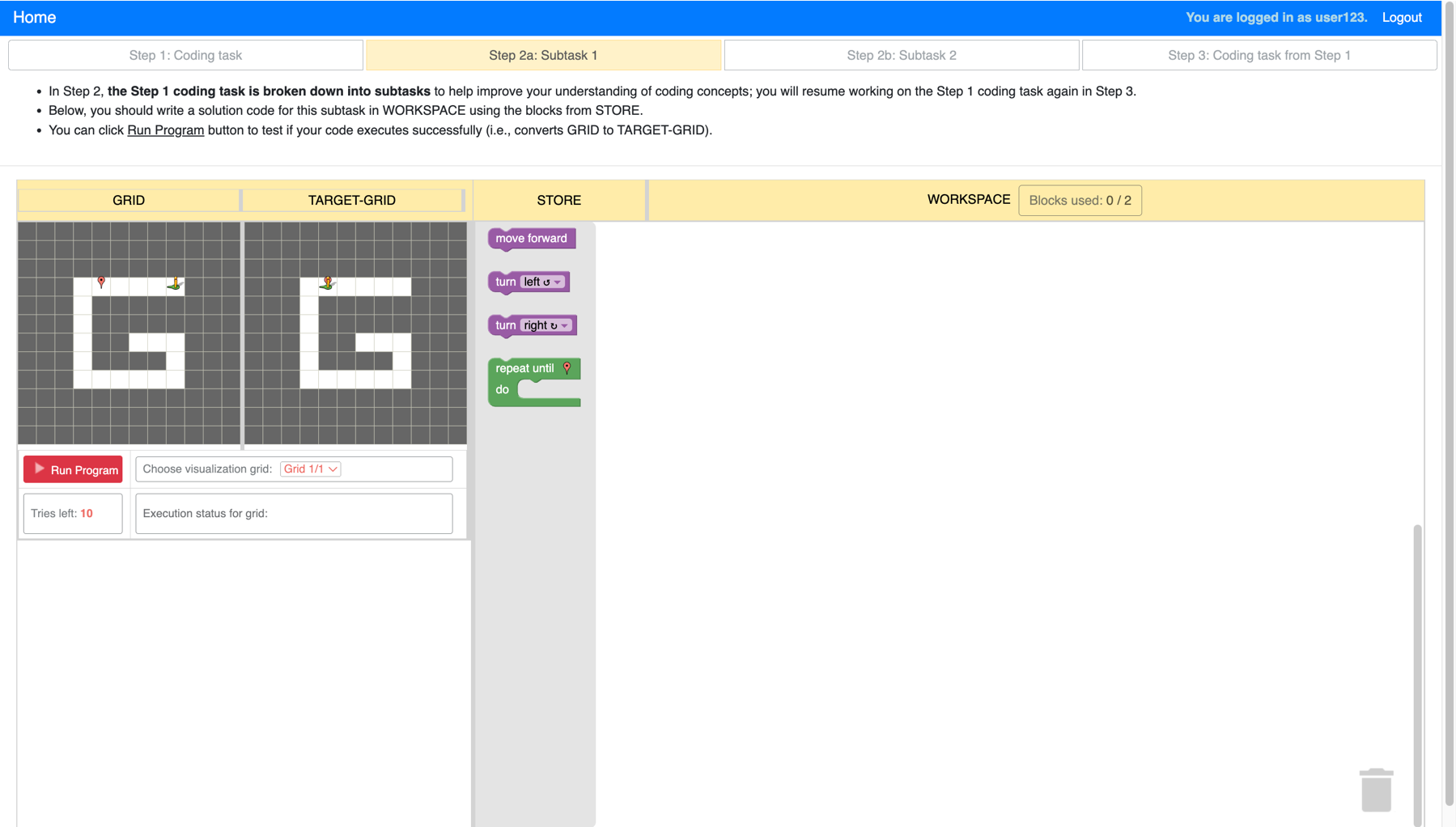}}
    \caption{Step $2$: Solve progression of subtasks}
    \end{subfigure}
    \hspace{3mm}
     \begin{subfigure}[t]{0.48\textwidth}
     \centering
        \frame{\includegraphics[trim={0.2cm 0.2cm 0.5cm 0.1cm},clip,height=4.6cm]{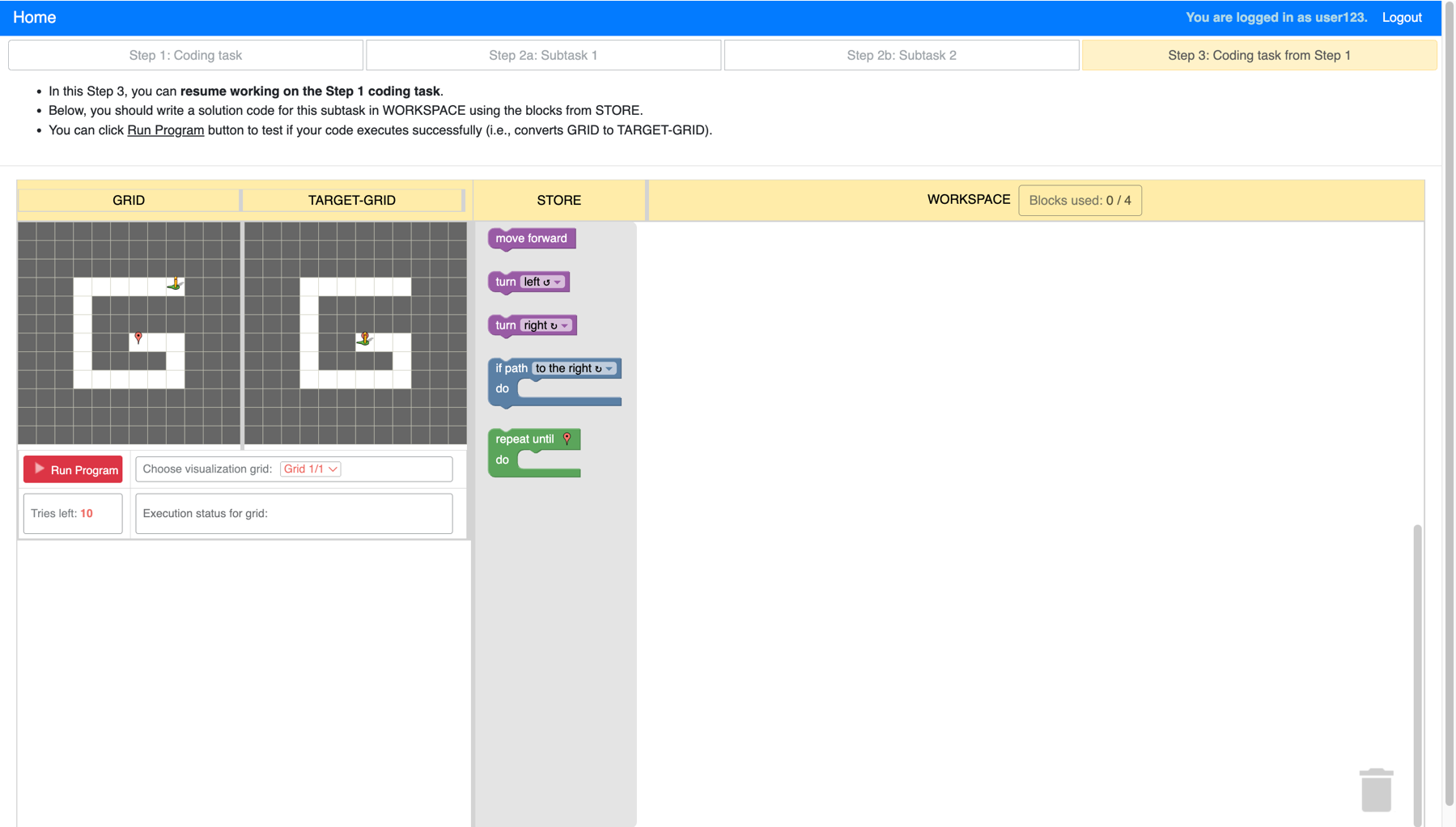}}
    \caption{Step $3$: Solve $\task^\textnormal{ref}$}
    \end{subfigure}
\caption{App Interface (See Appendix~\ref{appendix.sec.userstudy} and Footnote~\ref{footnote:appendix.userstudy})}
\label{fig:appendix.userstudy.app}
\end{figure*}

\subsection{App Interface: Additional Details}\label{appendix.userstudy.app.details}
In this section, we present additional details of our web app used for the user study.

We illustrate the three stages of our web app in Figure~\ref{fig:appendix.userstudy.app}.\footnote{We illustrate an updated version of our app. In this version, a user can solve the reference task in Step $1$. If successful, they exit the session. If unsuccessful, they proceed to Step $2$ and solve the progression of subtasks.\label{footnote:appendix.userstudy}} On the app, we have enabled our subtasking algorithm \synsubtasks{} and three block-based visual programming tasks: \hocd{}, \hocg{}, and \emph{Karel programming environment} based \textsc{Stairway}. A user can select one of these three tasks to practice on the platform, guided by a progression of subtasks synthesized by \synsubtasks.

%

}

\end{document}